\documentclass[twoside]{article}
\usepackage{cl}

\issue{16}{1}{1994}

\title{Anaphora and Discourse Structure}

\author{Bonnie Webber%
         \thanks{Division of Informatics, University of Edinburgh,
                 2 Buccleuch Place, Edinburgh UK EH8 9LW. E-mail:
                 bonnie.webber@ed.ac.uk}
       & Matthew Stone\\
         \affil{Edinburgh University} 
       & \affil{Rutgers University} \\ \
         \\
     Aravind Joshi & Alistair Knott \\
        \affil{University of Pennsylvania}      & \affil{University of Otago} \\
}

\runningtitle{Anaphora and Discourse Structure}
\runningauthor{Webber et al.}

\usepackage{epsfig}
\usepackage{amsmath}

\newcommand {\bc}{\begin{center}}
\newcommand {\ec}{ \end{center}}
\newcommand {\be} {\begin{enumerate}}
\newcommand {\ee} {\end{enumerate}}
\newcommand {\bd} {\begin{description}}
\newcommand {\ed} {\end{description}}

\usepackage{mylingmacros}

\begin{document}

\maketitle

\begin{abstract}
We argue in this paper that many common adverbial phrases generally
  taken to signal a discourse relation between syntactically connected units
  within discourse structure, instead work anaphorically to contribute
  relational meaning, with only indirect dependence on discourse
  structure. This allows
  a simpler discourse structure to provide scaffolding for compositional
  semantics, and reveals multiple ways in which the relational meaning
  conveyed by adverbial connectives can interact with that associated
  with discourse structure. We conclude by sketching out a lexicalised
  grammar for discourse that
  facilitates discourse interpretation as a product of compositional
  rules, anaphor resolution and inference.
\end{abstract}

\section*{Introduction}  

It is a truism that a text means more than the sum of its component
sentences. One source of additional meaning are {\em relations} taken
to hold between adjacent sentences ``syntactically'' connected
within a larger discourse structure.
However, it has been very difficult to say what discourse
relations there are, either theoretically \cite{mt88,kehler02,al:forth} or
empirically \cite{knott96}.

Knott's empirical attempt to identify and characterise
{\em cue phrases} as evidence for {\em discourse relations}
illustrates some of the difficulties. Knott used the following
theory-neutral test to identify cue phrases: For
a potential cue phrase $\phi$ in naturally occurring text, consider
in isolation the clause in
which it appears. If the clause appears incomplete without an adjacent
left context, while it appears complete if $\phi$ is removed, then
$\phi$ is a cue phrase. Knott's test produced a non-exhaustive list
of about 200 different phrases from 226 pages of text. He then
attempted to characterize the discourse relation(s) conveyed by
each phrase by identifying when (always, sometimes,
never) one phrase could substitute for another in a way that
preserved meaning. He then showed how these substitution patterns
could be a consequence of a set of semantic features and their
values. Roughly speaking, one cue phrase could {\em always} substitute for
another if it had the same set of features and values, {\em sometimes}
do so if it was less specific than the other in terms of its feature
values, and  {\em never} do so if their values conflicted for one or more
features. 

By assuming that cue phrases contribute meaning in a uniform way,
Knott was led to a set of surprisingly complex directed acyclic graphs
relating cue phrases in terms of features and their values, each graph
loosely corresponding to some family of discourse relations.  But what
if the relational meaning conveyed by cue phrases could in fact
interact with discourse meaning in multiple ways?  Then Knott's
substitution patterns among cue phrases may have reflected these
complex interactions, as well as the meanings of individual cue
phrases themselves. 

This paper argues that cue phrases {\em do} depend on  another
mechanism for conveying extra-sentential meaning  -- specifically,
{\em anaphora}.
One early hint that adverbial cue phrases (called here {\em discourse
connectives}) might be anaphoric can be found in an ACL workshop paper
in which Janyce Wiebe \shortcite{wiebe93} used the following example to
question the adequacy of tree structures for discourse.
\enumsentence{
a. The car was finally coming toward him.\\
b. He $[$Chee$]$ finished his diagnostic tests,\\
c. feeling relief.\\
d. {\em But then} the car started to turn right.
\label{wiebe}}
The problem she noted was that the discourse connectives {\em but} and
{\em then} appear to link clause (\ref{wiebe}d) to two different
things: ``then'' to clause (\ref{wiebe}b) in a {\em sequence} relation
-- i.e., the car starting to turn right being the next relevant event
after Chee's finishing his tests -- and ``but'' to a grouping of clauses
(\ref{wiebe}a) and (\ref{wiebe}c) -- i.e., reporting a contrast
between, on the one hand, Chee's attitude towards the car coming
towards him and his feeling of relief and, on the other hand, his
seeing the car turning right. (Wiebe doesn't give a name
to the relation she posits between (\ref{wiebe}d) and the grouping of
(\ref{wiebe}a) and (\ref{wiebe}c), but it appears to be some form of
{\em contrast}.) 

\begin{figure}[t]
\centerline{\epsfig{file=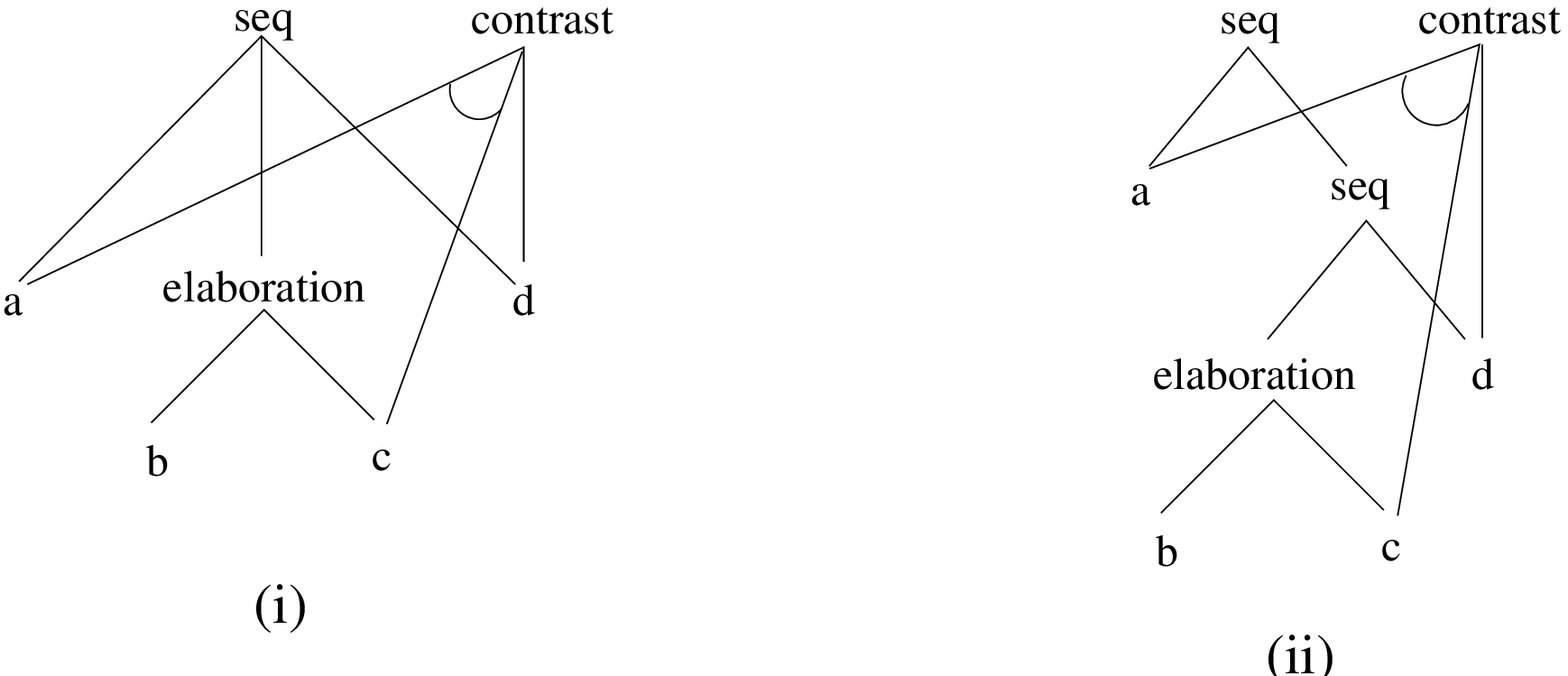,scale=0.6}}
\caption{Possible discourse structure for Example~\ref{wiebe}. Each root
and internal node is labelled by the type of relation that Wiebe takes
to hold between the daughters of that node. (i) uses an n-ary
branching {\em sequence} relation, while in (ii), {\em sequence} is
binary branching.}
\label{chee:fig}
\end{figure}

If these relations are taken to be the basis for discourse structure,
some possible discourse structures for this example are given
in Figure~\ref{chee:fig}. Such structures
might seem advantageous in 
allowing the semantics of the example to be computed directly by
compositional rules and defeasible inference. However, both
structures are directed acyclic graphs (DAGs), with acyclicity the
only constraint on what nodes can be connected. Viewed
syntactically, arbitrary DAGS are completely unconstrained 
systems. They substantially complicate interpretive rules for
discourse, in order for those rules to account for the relative scope
of unrelated operators and the contribution of syntactic nodes with
arbitrarily many parents.\footnote{A reviewer has suggested an
alternative analysis of (\ref{wiebe}) in which clause (\ref{wiebe}a) is
{\em elaborated} by clause (\ref{wiebe}b) which is in turn
{\em elaborated} by (\ref{wiebe}c), and clause (\ref{wiebe}d) stands in
both a {\em sequence} relation and a {\em contrast} relation with
the segment as a whole. While this might address Wiebe's problem, the
result is still a DAG, and such a fix will not address the
additional examples we present in Section~\ref{anaph:sec}, where
a purely structural account still requires DAGs with crossing arcs.}

\begin{figure}[h]
\centerline{\epsfig{file=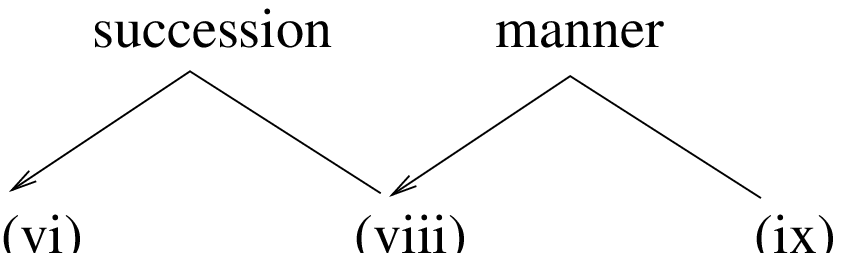,
    scale=.60}}
\caption{Simple multi-parent structure}
\label{zigzag:fig}
\end{figure}

We are not committed to trees as the limiting case of discourse
structure. For example, we agree, by and large, with the analysis that
\namecite{bateman99} gives of 
\enumsentence{
\ldots (vi) The first to do that were the German jewellers, (vii) in
particular Klaus Burie. (viii) And Morris followed very quickly after,
(ix) using a lacquetry technique to make the brooch, (x) and using
acrylics, (xi) and exploring the use of colour, (xii) and colour is
another thing that was new at that time \ldots}
in which clause (ix) stands in a {\em manner} relation with clause (viii),
which in turn stands in a {\em succession} (i.e., sequence) relation
with clause (vi). This is illustrated in Figure~\ref{zigzag:fig}. It
is a DAG (rather than a tree), but without crossing dependencies.

So it is the cost of moving to {\em arbitrary} DAGs for
discourse structure that we feel is too great to be taken 
lightly. This is what has led us to look for another explanation for
these and other examples of apparent complex and crossing dependencies
in discourse.

The position we argue for in this paper,
is that while adjacency and explicit conjunction (coordinating
conjunctions such as ``and'', ``or'', ``so'' and ``but'';
subordinating conjunction such as ``although'', ``whereas'',
``when'', etc.) imply discourse relations between (the interpretation of)
adjacent or conjoined discourse units, {\em discourse adverbials} such
as  ``then'', ``otherwise'', ``nevertheless'' and ``instead'' are {\em
  anaphors}, signalling a relation  between the interpretation of
their matrix clause and an entity in or derived from the discourse 
context. This position has four advantages.
\be
\item ~Understanding discourse adverbials as anaphors recognises their
{\em behavioral similarity} with the
pronouns and definite noun phrases (NPs) that are the
``bread and butter'' of previous work on anaphora. This is discussed
in Section~\ref{anaph:sec}. 
\item ~By understanding and exploring the full range of phenomena for
  which an anaphoric account is appropriate, we can better
characterise anaphors and devise more accurate algorithms for
resolving them. This is explored in Section~\ref{frame:sec}.
\item ~Any theory of discourse must still provide an account of how a
  sequence of adjacent discourse units (clauses, sentences, and the
  larger units that they can comprise) means more than
just the sum of its component units. This is a goal that researchers
have been pursuing for some time, using both {\em compositional} rules
and defeasible inference to determine these additional aspects of
meaning.\cite{al99,gardent97,hsam93,kehler02,pol-van96,scha88,schilder97b,schilder97a,vandenberg96}.
By factoring out that portion of discourse semantics that can be
handled by mechanisms already needed for resolving other forms of
anaphora and deixis, there is less need to stretch and possibly
distort compositional rules and defeasible inference to
handle everything.\footnote{There is an analogous situation at the
sentence level, where the relationship between syntactic structure and
compositional semantics is simplified by factoring away
inter-sentential anaphoric relations. Here the factorisation is so
obvious that one does not even think about any other possibility.}
Moreover, recognising the possibility of two separate relations (one
derived anaphorically and one associated with adjacency and/or a
structural connective) admits additional richness to discourse
semantics. Both points are discussed further in Section~\ref{inf:sec}.  

\item ~Understanding discourse adverbials as anaphors allows us to see
  more clearly how a lexicalised approach to the computation of
  clausal syntax and semantics extends naturally to the computation of
  discourse syntax and semantics, providing a single syntactic and
  semantic matrix
  with which to associate speaker intentions and other aspects of
  pragmatics. (Section~\ref{tag:sec}) 
\ee

The account we provide here is meant to be compatible with current
approaches to discourse semantics such as DRT \cite{kr93,ek97}, Dynamic
Semantics \cite{sg99}, and even SDRT \cite{asher:book,al:forth} --
understood as a representational scheme rather than an interpretive
mechanism. It is also meant to be compatible with more detailed analyses
of the meaning and use of individual discourse adverbials, such as
\cite{jr98a,jr98c,traugott95,traugott97}. 
It provides what we believe to be a more coherent account of {\em how}
discourse meaning is computed, rather than an alternative account of
{\em what} that meaning is or {\em what speaker intentions} it is being 
used to achieve. 

\section{Discourse Adverbials as Anaphors}
\label{anaph:sec}

\subsection{Discourse Adverbials do not behave like Structural
Connectives}
\label{no-cross:sec} 

We take the building blocks of the most basic level of
discourse structure to be explicit {\em structural connectives}
between adjacent discourse units (i.e., coordinating and subordinating
conjunctions, and ``paired'' conjunctions such as ``not 
only ... but also'', ``on the one hand ... on the other (hand)'',
etc.) and {\em  inferred relations} between adjacent discourse units
(in the absense of an explicit structural connective). Here,
adjacency is what triggers the inference. Consider the following
example: 
\enumsentence{You shouldn't trust John. He never returns what he
borrows.\label{trust-imp:ex}}
Adjacency leads the hearer to hypothesize that a discourse relation of
something like {\em explanation} holds between the two
clauses. Placing the subordinate conjunction (structural connective)
``because'' between the two clauses provides more evidence for this
relation. Our goal in this section is to convince the reader that many
{\em discourse adverbials} -- including ``then'', ``also'', ``otherwise'',
``nevertheless'', ``instead'' -- do not behave in this way.

Structural connectives and discourse adverbials do have one thing in
common: Like verbs, they can both be seen as heading a
predicate-argument construction; unlike verbs, their arguments are
independent clauses. For example, both the subordinate conjunction
``after'' and the adverbial ``then'' (in its temporal sense) can be
seen as binary predicates (e.g., {\em sequence}) whose
arguments are clausally-derived events, with the earlier event in
first position and the succeeding event in second.

\begin{figure}
\centerline{\epsfig{file=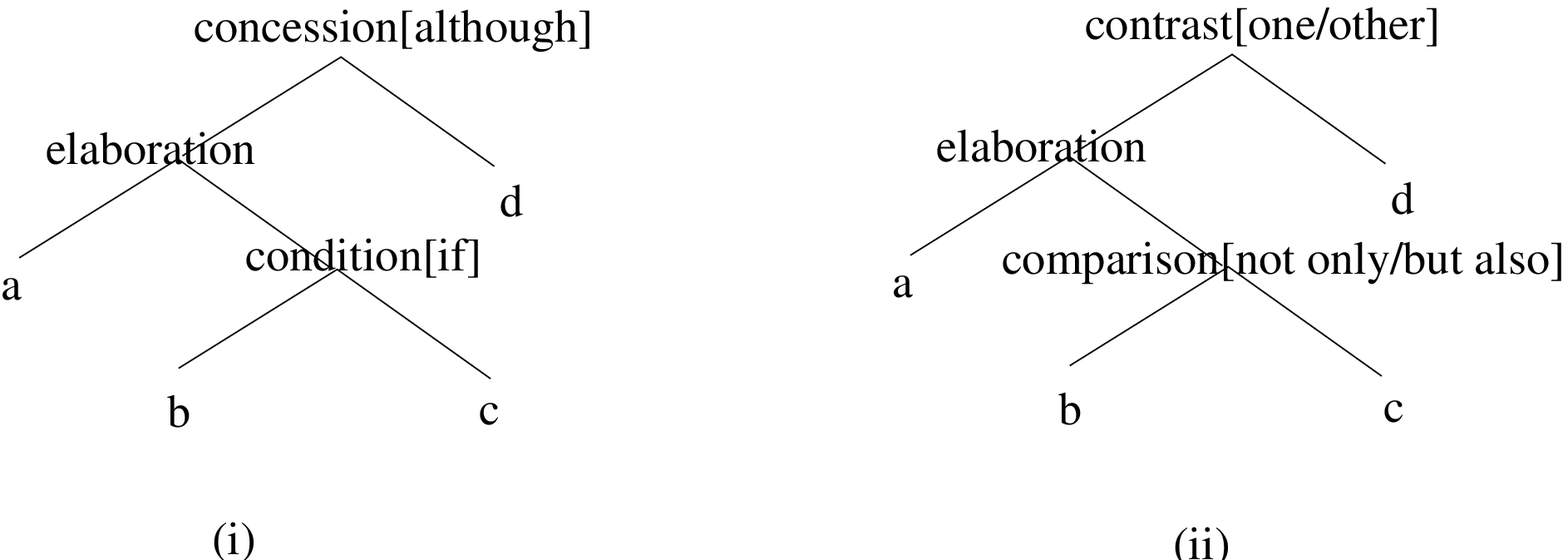,scale=0.6}}
\caption{Discourse structures associated with (i) Example~\ref{unx1:ex} 
and (ii) Example~\ref{unx2:ex}.}
\label{unx:fig}
\end{figure}

But that is the only thing that discourse adverbials and structural
connectives have in common. As we have pointed out in earlier papers
\cite{wkj99b,wksj99,wksj99:lorid}, structural connectives have two
relevant properties: (1) they admit {\em stretching} of
predicate-argument dependencies; and (2) they do not admit
{\em crossing} of those dependencies. This is most  obvious in the
case of preposed
subordinate conjunctions (Example~\ref{although:ex}) or ``paired''
coordinate conjunctions (Example~\ref{one-other:ex}). With such
connectives, the initial predicate signals that
its two arguments will follow.
\enumsentence{{\em Although} John is generous, he is hard to
       find.\label{although:ex}} 
\enumsentence{{\em On the one hand}, Fred likes beans. {\em On the
    other hand}, he's allergic to them.\label{one-other:ex}}
\noindent
Like verbs, structural connectives allow the distance
between the predicate and its arguments to be ``stretched'' over
embedded material, without loss of the dependency between them. For
the verb ``like'' and an object argument ``apples'', such stretching
without loss of dependency is illustrated in Example~\ref{stretch1:ex}.
\eenumsentence{
\item Apples John likes.
\item Apples Bill thinks he heard Fred say John
likes.\label{stretch1:ex}}
That this also happens with structural connectives and their
arguments, is illustrated in Example~\ref{unx1:ex} (in which the first
clause of Example~\ref{although:ex} is elaborated by another
preposed subordinate-main clause construction embedded within it)
and Example~\ref{unx2:ex} (in which the first conjunct of
Example~\ref{one-other:ex} is elaborated by another
paired-conjunction construction embedded within it).
Possible discourse structures for these examples are given in
Figure~\ref{unx:fig}.
\enumsentence{a. Although John is very generous -- \\
b. if you need some money,\\
c. you only have to ask him for it -- \\
d. he's very hard to find.\label{unx1:ex}} 
\enumsentence{a. On the one hand, Fred likes beans.\\
b. Not only does he eat them for dinner.\\
c. But he also eats them for breakfast and snacks.\\
d. On the other hand, he's allergic to them. \label{unx2:ex}}

\begin{figure}
\centerline{\epsfig{file=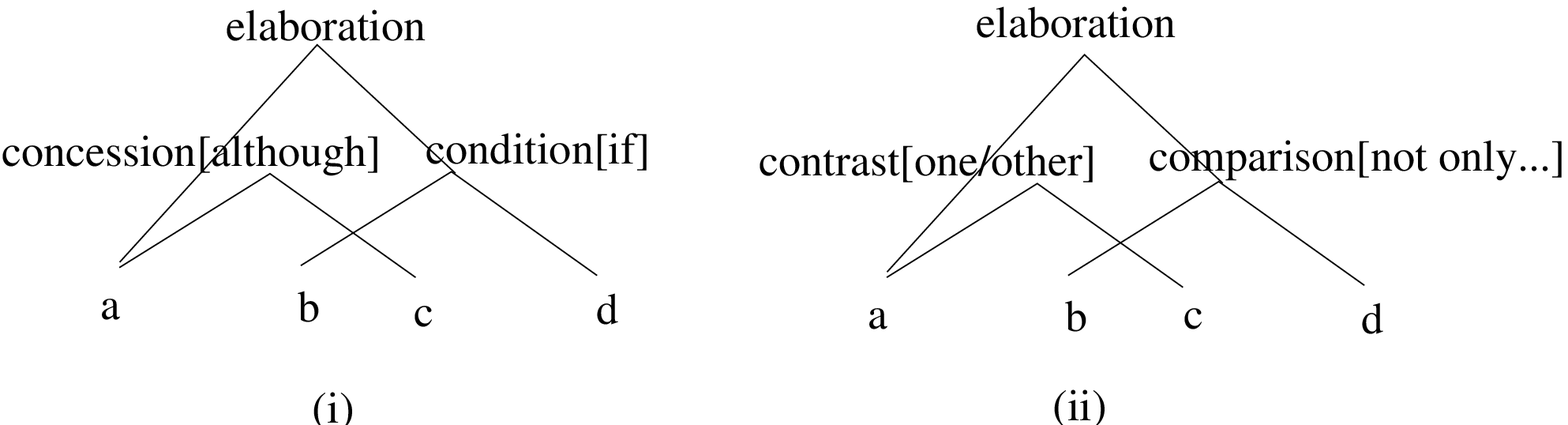,scale=0.6}}
\caption{(Impossible) discourse structures that would have to be
  associated with Example~\ref{x1:ex} (i) and with Example~\ref{x2:ex}
  (ii).}
\label{x:fig}
\end{figure}

\noindent
But, as already noted, structural connectives do not admit {\em
  crossing} of predicate-argument dependencies. If we do this with
Examples~\ref{unx1:ex} and ~\ref{unx2:ex}, we get
\enumsentence{a. Although John is very generous -- \\
b. if you need some money -- \\
c. he's very hard to find -- \\
d. you only have to ask him for it. \label{x1:ex}}  
\enumsentence{a. On the one hand, Fred likes beans.\\
b. Not only does he eat them for dinner.\\
c. On the other hand, he's allergic to them.\\
d. But he also eats them for breakfast and snacks.\label{x2:ex}}
Possible discourse structures for these (impossible) discourses are
given in Figure~\ref{x:fig}. 
Even if the reader finds no problem with these crossed versions,
they clearly do not mean the same thing as their uncrossed
counterparts: In (\ref{x2:ex}), ``but'' now appears to link
(\ref{x2:ex}d) with (\ref{x2:ex}c), conveying that despite being
allergic to  beans, Fred eats them for breakfast and snacks. And while
this might be inferred from (\ref{unx2:ex}), it is certainly not conveyed
directly.  As a consequence, we stipulate that structural connectives
do not admit crossing of their predicate-argument
dependencies.\footnote{A reviewer has asked how much ``stretching''
  is possible in discourse without losing its thread
  or having to rephrase later material in light of the intervening
  material. One could ask a similar question about the apparently
  unbounded dependencies of sentence-level syntax, which inattentive
  speakers are prone to lose track of and ``fracture''. Neither question
  seems answerable on theoretical grounds alone, demanding substantial amounts
  of empirical data from both written and spoken discourse. The point we
  are trying to make is simply that there is a difference in discourse
  between any amount of stretching and even the smallest amount of
  crossing.}

That is not all. Since we take the basic level of discourse structure to
be a consequence of (a) relations associated with explicit structural
connectives and (b) relations whose defeasible inference is triggered
by adjacency, we stipulate that {\em discourse
structure itself does not admit crossing structural dependencies}.
(In this sense, discourse structure may be truly simpler than sentence
structure. To verify this, one might examine the
discourse structure of languages such as Dutch that allow crossing
dependencies in sentence-level syntax. Initial cursory examination
does not give any evidence of crossing dependencies in Dutch discourse.)

If we now consider the corresponding properties of discourse
adverbials, we see that they do admit crossing of predicate-argument
dependencies, as shown in Examples~\ref{barolo:ex}-\ref{sickzoo:ex}.
\enumsentence{
a.~John loves Barolo.\\
b.~So he ordered three cases of the '97.\\
c.~But he had to cancel the order\\
d.~because {\em then} he discovered he was broke. \label{barolo:ex}}

\enumsentence{
a. High heels are fine for going to the theater.\\
b. But wear comfortable shoes\\
c. if {\em instead} you plan to go to the zoo.\label{shoezoo:ex}}

\enumsentence{
a. Because Fred is ill\\
b. you will have to stay home\\
c. whereas {\em otherwise} the two of you could have gone to the
zoo.\label{sickzoo:ex}} 

\begin{figure}
\epsfig{file=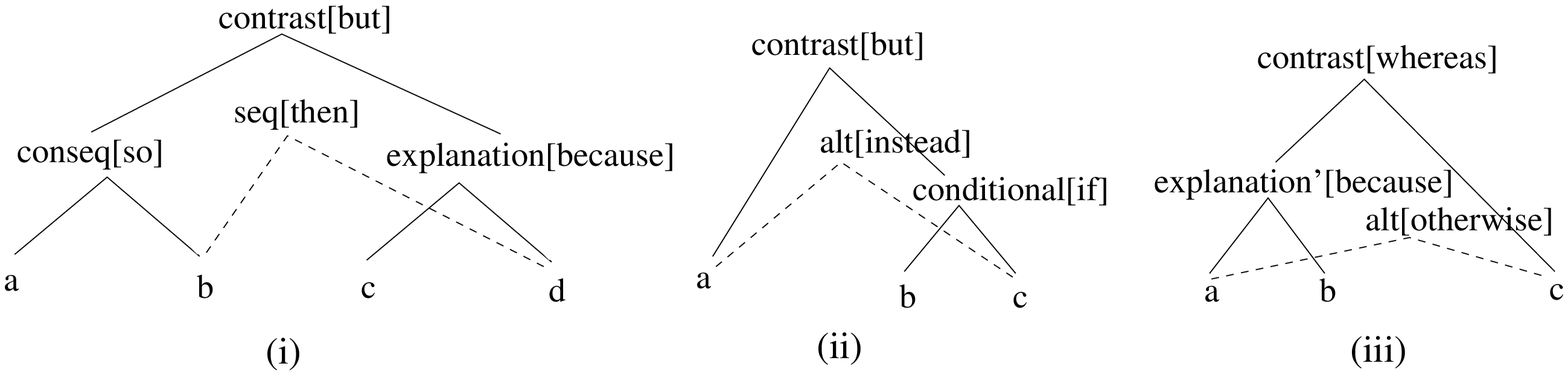,scale=0.6}
\caption{Discourse structures for
  Examples~\ref{barolo:ex}--\ref{sickzoo:ex}. Structural dependencies
  are indicated by solid lines and dependencies associate with
  discourse adverbials are indicated by dashed lines. ({\em
  explanation'} is the inverse of {\em explanation} -- i.e., with its
  arguments in reverse order. Such relations are used to maintain the
  given linear order of clauses.)}
\label{barolo:fig}
\end{figure}

Consider first the discourse  adverbial ``then'' in clause
(\ref{barolo:ex}d). For it to get its first argument from 
(\ref{barolo:ex}b) -- i.e., the event that the discovery in (d) is
``after'', it must cross the structural connection between clauses (c)
and (d) associated with ``because''. This crossing dependency is
illustrated in Figure~\ref{barolo:fig}i. Now consider the discourse
adverbial ``instead'' in clause (\ref{shoezoo:ex}c). For it to get its
first argument from (\ref{shoezoo:ex}a) -- i.e., going to the zoo
is an alternative to going to the theater -- it must 
cross the structural connection between clauses (\ref{shoezoo:ex}b) and
(\ref{shoezoo:ex}c) associated
with ``if''. This crossing dependency is illustrated in
Figure~\ref{barolo:fig}ii. Example~\ref{sickzoo:ex} is its mirror image:
For the  discourse adverbial
``otherwise'' in (\ref{sickzoo:ex}c) to get its first argument
from (\ref{sickzoo:ex}a) -- i.e., alternatives to the state/condition of
Fred being ill -- it must cross the structural connection
associated with ``because''. This is illustrated in
Figure~\ref{barolo:fig}iii.

Crossing dependencies are not unusual in discourse when one considers
anaphora (e.g., pronouns and definite NPs), as for example in: 
\enumsentence{Every man$_i$ tells every woman$_j$ he$_i$ meets that
  she$_j$ reminds him$_i$ of his$_i$ mother.}
\enumsentence{Sue$_i$ drives an Alfa Romeo. She$_i$ drives too fast.
Mary$_j$ races her$_i$ on weekends. She$_j$ often beats her$_i$.
\cite{strube98}}
This suggests that in Examples~\ref{barolo:ex}--\ref{sickzoo:ex}, the
relationship between  the discourse adverbial and its (initial) argument
from the previous discourse might usefully be taken to be anaphoric as
well.\footnote{We are aware that ``crossing'' examples such as
(\ref{barolo:ex})--(\ref{sickzoo:ex}) are rare in naturally--occurring
discourse. We believe that this is because they are only possible when,
as here,  strong constraints from the discourse adverbial
and from context prevent the adverbial from relating to the closest
(leftmost) eventuality or an eventuality coerced from that one.
But rarity doesn't necessarily mean ill-formedness or marginality,
as readers can see for themselves if they use Google to search the web
for strings such as ``because then'', ``if instead'', ``whereas
otherwise'', etc. and consider (a) whether the hundreds, even thousands,
of texts in which these strings occur are ill-formed, and (b) what
``then'', ``instead'' and ``otherwise'' are relating in these texts.
One must look at rare events if one is studying complex linguistic
phenomena in detail. Thus it is not the case that only common
things in language are real or worth further study.}

\subsection{Discourse Adverbials do behave like Anaphors} 
\label{moresim:sec}

There is additional evidence to suggest that ``otherwise'', ``then''
and other discourse adverbials are anaphors. First, anaphors in the
form of definite and demonstrative NPs can take implicit
material as their referents. For example, in 
\enumsentence{Stack five blocks on top of one another. Now close your
eyes and try knocking \{{\em the tower}, {\em this
  tower}\} over with your nose.} 
both NPs refer to the structure which is the implicit
result of the block stacking. (Further discussion of such examples
can be found in \cite{isard75,dale92,wb92}.)
The same is true of discourse adverbials. In
\enumsentence{Do you want an apple? {\em Otherwise} you can have a
   pear.\label{pear:ex}}
the situation in which you can have a pear is one in which you don't
want an apple -- i.e., where your answer to the question is
``no''. But this answer isn't there structurally: it is only
inferred. While it appears natural to resolve an 
anaphor to an inferred entity, it would be much more difficult to
establish such links through purely structural connections: to do so
would require complex transformations that introduce
invisible elements into discourse syntax with no deeper motivation.
For example, in (\ref{pear:ex}), we would need a rule that takes a
discourse unit consisting solely of a yes/no question $P?$ and replaces
it with a complex segment consisting of $P?$ and the clause
\emph{it is possible that P}, with the two related by something like
{\em elaboration}.  Then and only then could
we account for the interpretation of the subsequent \emph{otherwise}
structurally, by a syntactic link to the covert material (i.e., to
the possibility that $P$ holds, which \emph{otherwise} introduces
an alterative to).

Secondly, discourse adverbials have a wider range of options with
respect to their initial argument than do structural connectives
(i.e., coordinating and subordinating conjunctions). The latter are
constrained to linking a discourse unit on the {\em right frontier} of
the evolving discourse (i.e., the clause, sentence and larger
discourse units to its immediate left). Discourse adverbials are not
so constrained. To see this, consider the following example:
\enumsentence{\label{arg2a:ex}
If the light is red, stop. {\em Otherwise} you'll get a ticket.\\
({\em If you do something other than stop, you'll get a ticket.})}
This can be paraphrased using the conjunction ``or''
\begin{quote}
If the light is red, stop, or you'll get a ticket.
\end{quote}
Here ``or'' links its right argument to a unit on the {\em right
  frontier} of the evolving discourse -- in this case,  the clause
  ``stop'' on its immediate left. Now consider the related 
example
\enumsentence{\label{arg2b:ex}
If the light is red, stop. {\em Otherwise} go straight on.\\
({\em If the light is not red, go straight on.})}
This cannot be paraphrased with ``or'', as in
\enumsentence{If the light is red, stop, or go straight
  on. \label{stopor2:ex}}
even though both ``stop'' and ``if the light is red,
stop'' are on the right frontier of the evolving discourse
structure. This is because ``otherwise'' is accessing something else,
so that (\ref{stopor2:ex}) means something quite different from 
either (\ref{arg2a:ex}) or (\ref{arg2b:ex}). What ``otherwise'' is
accessing, which ``or'' cannot, is the interpretation of the condition
alone.\footnote{This was independently pointed out by several people
when this work was presented at ESSLLI'01 in Helsinki, August 2001.
The authors would like to thank Natalia Modjeska, Lauri Karttunen,
Mark Steedman, Robin Cooper and David Traum for bringing it to their
attention.} Thus discourse
adverbials, like other anaphors, have access to material that is not
available to structural connectives.

Finally, discourse adverbials, like other anaphors, may require
semantic representations in which their arguments are {\em bound variables}
ranging over discourse entities.  That is, while it might
be possible to represent ``Although $P$, $Q$'' using a binary modal operator
\enumsentence{\emph{although(p, q)}}
where formulas $p$ and $q$ translate the sentences $P$ and $Q$
that ``although'' combines, we cannot represent ``$P$ ... Nevertheless, $Q$''
this way. We need something more like
\enumsentence{\emph{p $\wedge$ nevertheless(e, q)}\label{var:ex}}
The motivation for the variable $e$ in this representation is
that discourse adverbials, like pronouns, can appear intra-sententially
in an analogue of donkey sentences.  Donkey sentences such as
Example~\ref{donkey:ex} are a special kind of bound-variable reading.    
\enumsentence{Every farmer who owns a donkey feeds {\em it}
rutabagas. \label{donkey:ex}} 
In donkey sentences, anaphors are interpreted as co-varying with their
antecedents: the \emph{it} that is being fed in (\ref{donkey:ex})
varies with the
farmer who feeds it.  However, these anaphors appear in a structural
and interpretive environment in which a direct syntactic relationship
between anaphor and antecedent is normally impossible, so cannot be
a reflex of true binding in the
syntax-semantics interface.  Rather, donkey sentences show that
\emph{discourse} semantics has to provide variables to translate
pronouns, and that \emph{discourse} mechanisms must interpret these
variables as bound---even though the pronouns appear ``free'' by
syntactic criteria.

Thus, it is significant that discourse adverbials can appear in their
own version of donkey sentences, as in 
\eenumsentence{\label{donkeyd:ex}
\item Anyone who has developed innovative new software, has {\em then} had
  to hire a laywer to protect his/her interests. (i.e., {\em after
  developing innovative new software})
\item Several people who have developed innovative new software, have
  {\em nevertheless} failed to profit from it. (i.e.,
  {\em despite having developed innovative new software})
\item Every person selling ``The Big Issue'' might {\em otherwise}
  be asking for spare change. (i.e., {\em if s/he weren't selling
``The Big Issue''})}
The examples in (\ref{donkeyd:ex}) involve binding in the interpretation of
discourse adverbials.  In (\ref{donkeyd:ex}a), the temporal use of
\emph{then} locates each hiring event after the \emph{corresponding}
software-development.  Likewise in (\ref{donkeyd:ex}b), the adversative
use of \emph{nevertheless} signals each developer's eye-opener in failing
to turn the \emph{corresponding} profit.  And in (\ref{donkeyd:ex}c),
\emph{otherwise} envisions each person begging if \emph{that person}
weren't selling "The Big Issue".

Such bound interpretations require variables in
the semantic representations and alternative values for them in some
model -- hence the representation given in (\ref{var:ex}).  Indeed,
it is clear that the binding
here has to be the discourse kind, not the syntactic kind -- for the
same reason as in (\ref{donkey:ex}), although we cannot imagine anyone arguing
otherwise, since discourse adverbials have always been treated as
elements of discourse interpretation.  So the variables must
be the discourse variables usually used to translate other kinds of
discourse anaphors.\footnote{While Rhetorical Structure Theory (RST)
  \cite{mt88} was developed as an account of the relation between
  adjacent units within a text, Marcu's guide to RST annotation
  \cite{marcu99} has added an ``embedded'' version of each RST
  relation in order to handle examples such as (\ref{donkeyd:ex}).
  While this importantly recognises that material in an embedded
  clause can bear a semantic relation to
  its matrix clause, it does not contribute to understanding the nature of
  the phenomenon.}

These arguments have been directed at the behavioral similarity
between discourse adverbials and what we normally take
to be discourse anaphors. But this isn't the only reason to recognise
them as anaphors: In the next section, we suggest a framework for
anaphora that is sufficiently broad enough to include discourse
adverbials as well as definite and demonstrative pronouns and NPs, and
other discourse phenomena that have been argued to be anaphoric, such
as VP ellipsis \cite{hardt92,kehler02}, tense \cite{partee84,webb88}
and modality \cite{kibble95,frank/kamp:97,stoneHardt99}. 

\section{A Framework for Anaphora}
\label{frame:sec}

Here we show how only a single extension to a general framework for
discourse anaphora is needed to cover discourse adverbials. The
general framework is presented in Section~\ref{over:sec}, and the
extension in Section~\ref{relanaph:sec}.

\subsection{Discourse referents and anaphor interpretation}
\label{over:sec}

The simplest discourse anaphors are {\em coreferential} -- definite
pronouns and definite NPs that denote one (or more) discourse
referents in {\em focus} within the current
discourse context. (Under {\em coreference} we include {\em split
  reference}, where a plural anaphor such as ``the companies'' denotes
all the separately mentioned companies in focus within the discourse
context.)  Much has been written about the factors affecting
what discourse referents are taken to be in focus. For a recent
review by Andrew Kehler, see Chapter~18 of \cite{jurafsky00}. For the
effect of different types of quantifiers on discourse referents and
focus, see \cite{kibble95}. 

Somewhat more complex than {\em coreference} is
{\em indirect anaphora} \cite{indiana} (also called {\em partial anaphora}
\cite{luperfoy92}, {\em textual ellipsis} \cite{hms96}, {\em
  associative anaphora} \cite{cosse96}, {\em bridging anaphora}
\cite{clark75,cm81,not99}, and {\em inferrables} \cite{prince92}), where
the anaphor -- usually a definite NP -- denotes 
a discourse referent {\em associated with} one (or more) discourse
referents in the current discourse context -- e.g.,
\enumsentence{Myra darted to {\em a phone} and picked up
{\em the receiver}.\label{myra:ex}}
Here {\em the receiver} denotes the receiver associated with (by
virtue of being {\em part of}) the already-mentioned phone Myra darted
to. 

Coreference and indirect anaphora can be uniformly modelled by saying
that the discourse referent $e_{\alpha}$ denoted by an anaphoric
expression $\alpha$ is either equal to or associated with
an existing discourse referent $e_r$ -- that is, $e_{\alpha}$=$e_r$ or
$e_{\alpha}\in$assoc($e_r$).

But coreference and associative anaphora do not exhaust the space of
constructs that derive all or part of their sense from the discourse
context and are thus anaphoric. Consider ``other NPs''
\cite{bierner01,bw00,modjeska01,modjeska02}, as in:
\enumsentence{Sue grabbed one phone, as Tom darted to
    {\em the other phone}.\label{tom:ex}}
While ``other NPs'' are clearly anaphoric, should the referent of
``the other phone'' ($e_\alpha$) -- the phone 
other than the one Sue grabbed ($e_r$) -- be simply considered a case
of $e_\alpha\in$assoc($e_r$)?  Here are two reasons why not.

First, in all cases of associative anaphora discussed in the literature,
possible associations have depended only on the antecedent $e_r$ and
not on the anaphor. For example, only antecedents that have 
parts participate in whole-part associations
(e.g. phone~$\rightarrow$~receiver). Only antecedents with functional
schemata participate in schema-based associations
(e.g. lock~$\rightarrow$~key). In (\ref{tom:ex}), 
the relationship between $e_\alpha$, the referent of ``the other
phone'', and its antecedent, $e_r$, depends in part on the anaphor, and
not just on the antecedent -- in particular, on the presence of
the word ``other''. Secondly, we also have examples such as
\enumsentence{Sue lifted the receiver as Tom darted to {\em the
    other phone}.\footnote{Modjeska \shortcite{modjeska01} discovered
    such examples in the British National Corpus.}\label{phone:ex}}
where the referent of ``the other phone'' ($e_\alpha$) is
the phone {\em other than} the phone {\em associated with} the
receiver that Sue lifted. Together, these two points argue for a third possibility, in
which an anaphoric element can convey a specific function $f_{\alpha}$
that is idiosyncratic to the anaphor, which may be applied to either
$e_r$ or an associate of $e_r$. The result of that application is
$e_\alpha$. For want of a better name, we will call these
{\em lexically-specified} anaphors. 

Other lexically-specified anaphors include noun phrases headed by
``other'' (Example~\ref{dog:ex}),  NPs with ``such'' but no
post-modifying ``as'' phrase (Example~\ref{lobster:ex}), comparative
NPs with no post-modifying ``than'' phrase (Example~\ref{dogs:ex}),
and the pronoun ``elsewhere'' (Example~\ref{else:ex}) \cite{bierner:phd}.
\enumsentence{Some dogs are constantly on the move. {\em Others} lie
around until you call them.\label{dog:ex}}

\enumsentence{I saw a 2kg lobster in the fish store yesterday. The
  fishmonger said it takes about 5 years to grow to {\em such a size}.
  \label{lobster:ex}}

\enumsentence{Terriers are very nervous. {\em Larger dogs} tend to have
  calmer dispositions.\label{dogs:ex}}

\enumsentence{I don't like sitting in this room. Can we move {\em
    elsewhere}?\label{else:ex} }

To summarize the situation with anaphors so far, we have {\em
coreference} when $e_{\alpha}$=$e_r$, {\em indirect anaphora} 
when $e_{\alpha}\in$assoc($e_r$), and {\em lexically-specified
anaphora} when $e_{\alpha}$=$f_{\alpha}(e_i)$ where
$e_i$=$e_r$ or $e_i\in$assoc($e_r$).

\subsection{Discourse Adverbials as Lexical Anaphors}
\label{relanaph:sec}

There is nothing in this generalised approach to discourse anaphora
that requires that the source of $e_r$ be an NP, or that anaphor be a
pronoun or NP. For example, the antecedent $e_r$ of a singular
demonstrative pronoun (in English, ``this'' or ``that'') is often an
{\em eventuality} that derives from a clause, a sentence, or a larger
unit in the recent discourse
\cite{asher:book,byron02,eckert01,webb91}. We will see that this is
the case with discourse adverbials as well.

The extension we make to the general framework presented above in
order to include discourse adverbials as discourse anaphors, is to
allow more general functions $f_\alpha$ to be associated with
lexically-specified anaphors. In particular, for the discourse
adverbials considered in this paper, the function associated with an
adverbial maps its anaphoric argument -- an {\em eventuality} derived
from the current discourse context -- to a function that
applies to the interpretation of the adverbial's matrix clause (itself
an eventuality). The result is a {\em binary relation} that holds
between the two eventualities and is added to the discourse context. 
For example,  in
\enumsentence{John loves Barolo. So he ordered three cases of the
  '97. But he had to cancel the order because he {\em then}
discovered he was broke. \label{then1:ex}}
``then'', roughly speaking, contributes the fact that its
matrix clause event (John finding he was broke) is
{\em after} the anaphorically-derived event of his ordering the
wine.\footnote{Words and phrases that function as discourse
adverbials usually have other roles as well -- e.g., ``otherwise''
also serves as an adjectival modifier, in ``I was {\em otherwise}
occupied with grading exams''. This overloading of closed-class
lexico-syntactic items is not unusual in English, and any ambiguities
that arise must be handled as part of the normal ambiguity resolution
process.} Similarly, in
\enumsentence{John didn't have enough money to buy a mango.
{\em Instead}, he bought a guava. \label{guava:ex}}
``instead'' contributes the fact that its matrix clause event
(buying a guava) is as an alternative to the anaphorically derived
event of buying a mango. The relation between the two sentences is
something like {\em result}, as in ``So instead, he bought a
guava.'' 

Note that our only concern here is with the compositional and anaphoric
{\em mechanisms} by which adverbials contribute meaning. For
detailed analysis of their lexical semantics (but no attention to
mechanism), the reader is referred to
\cite{jr98a,jr98c,lagerwerf98,traugott95,traugott97} and others.

Formally, we represent the function that a discourse adverbial
$\alpha$ contributes, as a $\lambda$-expression involving a binary
relation $R_{\alpha}$ that is idiosyncratic to $\alpha$, one of whose
arguments (represented here by the variable $EV$) is resolved
anaphorically: 
\begin{quote}
$\lambda x\ .\ R_{\alpha}(x,EV)$
\end{quote}
$R_{\alpha}$ gets its other argument compositionally, when this 
$\lambda$-expression is applied to $\alpha$'s matrix clause $S$
interpreted as an eventuality $\sigma$ -- that is,
\begin{quote}
$[\lambda x\ .\ R_{\alpha}(x,EV)] \sigma \equiv R_{\alpha}(\sigma,EV)$
\end{quote}
The result of both  function application and resolving $EV$ to some
eventuality $e_i$ derived from the discourse context either
directly or by association, is the proposition $R_{\alpha}(\sigma,e_i)$,
one of whose arguments ($e_i$) has been supplied by the discourse
context and the other ($\sigma$) has been supplied compositionally
from syntax. 

Note that this is a {\em formal} model, meant to have no implications
for how {\em processing} takes place. Our view of processing is that
it is triggered by the discourse adverbial and its matrix clause.
Given $\alpha$ and $\sigma$, the resolution process
finds an eventuality $e_i$ (or creates an appropriate one by a
bridging inference, as illustrated in the next section) such that
$R_{\alpha}(\sigma,e_i)$ makes sense with respect to the
discourse. This is best seen as a constraint satisfaction problem
similar to that of resolving a {\em discourse deictic}
\cite{asher:book,byron02,eckert01,webb91}. 
That is, the process involves finding or deriving an eventuality
from the current discourse context, that meets the constraints of
the adverbial with respect to the eventuality interpretation of the
matrix clause. (Examples of this are given throughout the rest of the
paper.)

\subsection{A Logical Form for Eventualities}
\label{gen-anaph:sec}

Before using this generalised view of anaphora to show what discourse
adverbials contribute to discourse and how they interact with
discourse relations that arise from adjancency or explicit discourse
connectives, we briefly describe how we represent clausal
interpretations in logical form (LF). 

Essentially, we follow \cite{hobbs85} in using a rich ontology and a
representation scheme that makes explicit all the individuals and
{\em abstract objects} (i.e., propositions, facts/beliefs and
eventualities) \cite{asher:book} involved in the logical form (LF)
interpretation of an utterance. We do so because we want to make
intuitions about individuals, eventualities, lexical meaning and
anaphora as clear as possible. But certainly, other forms of
representation are possible. 

In this LF representation scheme, each clause and each
relation between clauses is indexed by the label of its associated
{\em abstract object}. So, for example, the LF interpretation of the
sentence
\enumsentence{John left because Mary left.}
would be written
\begin{quote}
$e_1$:left(j) $\wedge$ john(j) $\wedge$ $e_2$:left(m)
$\wedge$ mary(m) $\wedge$ $e_3$:because($e_1$,$e_2$)
\end{quote}
where the first argument of the asymmetric binary predicate {\em
because} is the consequent and the second is the eventuality leading
to this consequent. Thus when ``because'' occurs sentence-medially, as
in the above example, the eventuality arguments are in the same order
as their corresponding clauses occur in the text. When ``because''
occurs sentence-initially (as in ``Because Mary left, John did''), the
interpretation of the second clause (``John $[$left$]$) will appear as
the first argument and the interpretation of the first clause (``Mary
left'') will appear as the second.\footnote{We are not claiming to
  give a detailed semantics of discourse connectives except insofar as
  they may affect how discourse adverbials are resolved. Thus, for
  example, we are not bothering to distinguish between different
  senses of ``because'' (epistemic vs. non-epistemic), ``while''
  (temporal vs. concessive), ``since'' (temporal vs. causal), etc. Of
  course, these distinctions are important to discourse
  interpretation, but they are independent of and orthogonal to the
  points made in this paper.  Similarly, \namecite{asher:book} argues
  that a simple ontology of eventualities is too coarse-grained, and
  that discourse representations need to distinguish different kinds
  of abstract objects, including actions, propositions and facts as
  well as eventualities.  Different discourse connectives will require
  different kinds of abstract objects as arguments.  This distinction
  is also orthogonal to the points made in this paper, because we can
  understand these abstract referents to be associates of the
  corresponding Hobbsian eventualities and leave the appropriate
  choice to the lexical semantics of discourse
  connectives. \namecite{byron02} advocates a similar approach to
  resolving discourse anaphora.}

The set of available discourse referents includes both individuals
like $j$ and $m$, but also abstract objects like $e_1$ and $e_2$.
We then represent resolved anaphors by re-using these discourse
referents. So, for example, the LF interpretation of the follow-on
sentence 
\enumsentence{This upset Sue.}
would be written
\begin{quote}
$e_4$:upset'(DPRO, s) $\wedge$ sue'(s)
\end{quote}
where DPRO is the anaphoric variable contributed by the demonstrative
pronoun ``this''. Since the 
subject of ``upset'' could be either the eventuality of John's leaving
or the fact that he left because Mary left, DPRO could be resolved
to either $e_1$ or $e_3$ -- i.e, 
\begin{quote}
a. $e_4$:upset'($e_1$, s) $\wedge$ sue'(s)\\
b. $e_4$:upset'($e_3$, s) $\wedge$ sue'(s)
\end{quote}
depending on whether one took Sue to have been upset by (a) 
John's leaving or (b) that he left because Mary left.

\subsection{The Contribution of Discourse Adverbials to Discourse Semantics}
\label{semcontrib:sec}

Here we step through some examples of discourse adverbials and how
they make their semantic contribution to the discourse context. We
start with Example~\ref{then1:ex}, repeated here as (\ref{barolo-rep:ex}).
\enumsentence{a. John loves Barolo.\\
b. So he ordered three cases of the '97.\\
c. But he had to cancel the order\\
d.  because he {\em then} discovered he was broke. \label{barolo-rep:ex}}
Using the above LF representation scheme and our notation from
Section~\ref{relanaph:sec}, namely
\begin{itemize}
\addtolength{\itemsep}{-4pt}
\item $\alpha$ = the anaphoric expression (here, the discourse adverbial)
\item $R_{\alpha}$ = the relation name linked with $\alpha$
\item $S$ = the matrix clause/sentence containing $\alpha$
\item $\sigma$ = the interpretation of $S$ as an abstract object
\end{itemize}
and ignoring, for now, the conjunction ``because'' (to be
discussed in Section~\ref{inf:sec}), the relevant elements of
(\ref{barolo-rep:ex}d) can be represented as: 
\begin{quote}
$\alpha$ = {\em then}\\
$R_{\alpha}$ = {\em after}\\ 
$S$ = he $[$John$]$ discovered he was broke\\
$\sigma$ = $e_4$:find(j,$e_5$), where $e_5$:broke(j)
\end{quote}
This means that the unresolved interpretation of
(\ref{barolo-rep:ex}d) is 
\begin{quote}
$[\lambda$x . $R_{\alpha}$(x,$EV$)$] \sigma \ \equiv \ [\lambda$x
. {\em after}(x,$EV$)$] e_4$ $\equiv$  {\em after}($e_4,EV$)
\end{quote}
The anaphoric argument $EV$ is resolved to the eventuality $e_2$,
derived from (\ref{barolo-rep:ex}b) -- $e_2$:order(j, $c_1$).
\begin{quote}
{\em after}($e_4$,$EV$) $\rightarrow$ {\em after}($e_4$,$e_2$)
\end{quote}
That is, the eventuality of John finding he was broke is after
that of John ordering three cases of the '97 Barolo.
The resulting proposition {\em after}($e_4$,$e_2$) would be given its
own index, $e_6$, and added to the discourse context. 

When ``then'' it understood temporally, as it is above, as opposed to
logically, it requires a culminated eventuality from the discourse
context as its first argument (which \cite{vendler67} calls an
{\em achievement} or an {\em accomplishment}). The ordering event in
(\ref{barolo-rep:ex}b) is such an Vendlerian accomplishment. In
Example~\ref{then2:ex} though, there is no culminated eventuality in
the discourse context for ``then'' to take as its first argument.
\enumsentence{a. Go west on Lancaster Avenue.\\
b. {\em Then} turn right on County Line. \label{then2:ex}}
How does (\ref{then2:ex}b) get its interpretation?

As with (\ref{barolo-rep:ex}d), the relevant elements of
(\ref{then2:ex}b) can be represented as
\begin{quote}
$\alpha$ = {\em then}\\
 $R_{\alpha}$ = {\em after}\\ 
$S$ = turn right on County Line\\
$\sigma$ = $e_3$:turn-right($you$, $county\_line$)
\end{quote}
and the unresolved interpretation of (\ref{then2:ex}b) is thus
\begin{quote}
$[\lambda$ x . {\em after}(x, $EV$)$] e_3$ $\equiv$
    {\em after}($e_3$, $EV$)
\end{quote}
As for resolving $EV$, in a well-known paper, \namecite{ms88} discuss
several ways in which an  eventuality of one type (e.g., a {\em process})
can be coerced into an eventuality of another type (e.g., an {\em
 accomplishment}, which Moens and Steedman call a {\em culminated
process}). In this 
case, the matrix argument of ``then'' (the eventuality of ``turning
right on County Line'') can be used to coerce the process eventuality
in (\ref{then2:ex}b) into a culminated process of ``going west on
Lancaster Avenue until County Line''. We treat this coercion as a type
of {\em associative} or {\em bridging} inference, as in the examples
discussed in Section~\ref{over:sec}.  That is,
\begin{quote}
$e_2$ = culmination($e_1$)$\in$assoc($e_1$),  where $e_1$:go-west($you$,
$lancaster\_ave$) 
\end{quote}
Taking this $e_2$ as the anaphoric argument $EV$ of ``then'' yields
the proposition
\begin{quote}
{\em after}($e_3$, $e_2$)\\
\end{quote}
That is, the eventuality of turning right onto City Line is after that
of going west on Lancaster Avenue to City Line. This proposition
would be indexed and added to the discourse context. 

It is important to stress here that the level of representation we are
concerned with is essentially a {\em logical form} (LF) for
discourse. Any reasoning that might then have
to be done on their content might then require making explicit the
different modal and temporal contexts involved, their accessibility
relations, the status of abstract objects as facts, propositions or
eventualities, etc. But as our goal here is primarily to capture the
{\em mechanism} in which discourse adverbials are involved in discourse
structure and discourse semantics, we will continue to assume for as
long as possible that a LF representation will suffice.

Now it may appear as if there is no difference between treating
adverbials as anaphors and treating them as structural connectives,
especially in cases like (\ref{then2:ex}) where the antecedent comes
from the immediately left-adjacent context, and where the only
obvious semantic relation between the adjacent sentences appears to be
the one expressed by the discourse adverbial. (Of course, there may
also be a separate {\em intentional} relation between the two
sentences \cite{mp92}, independent of the relation conveyed by the
discourse adverbial.)

One must distinguish, however, between whether a theory allows a
distinction to be made and whether that distinction needs to be made in a
particular case. It is clear that there are many examples where
the two approaches (i.e., a purely structural treatment of all
connectives, versus one that treats adverbials as linking into the
discourse context anaphorically) appear to make the same prediction.
However, we have already demonstrated cases where a purely structural
account makes the wrong prediction, and in the next section, we will
demonstrate the additional power of an account that allows for {\em two}
relations between an adverbial's matrix clause or sentence and the
previous discourse -- one arising from the anaphoric connection and
the other inferred from adjacency or conveyed explicitly by a
structural connective.

Before closing this section, we want to step through Examples
\ref{arg2b:ex}--\ref{stopor2:ex}, repeated here as
Examples~\ref{red1:ex}--\ref{red2:ex}.
\enumsentence{If the light is red, stop. {\em Otherwise} you'll get a
  ticket.\label{red1:ex}}
\enumsentence{If the light is red, stop. {\em Otherwise} go straight
  on.\label{red2:ex}}
Roughly speaking, ``otherwise'' conveys that the
\emph{complement} of its anaphorically-derived argument
serves as the \emph{condition} under which the interpretation of
its structural argument holds. (This complement must be with respect to
some contextually relevant set.\footnote{\namecite{kkw01} demonstrate that 
the {\em Information Structure} of sentences in the previous discourse
(theme-rheme partitioning, as well as focus within theme and within
rheme \cite{steed01:LI}) can influence what eventualities  
$e_r$ are available for
resolving the anaphorically derived argument of ``otherwise''. This
then correctly predicts different interpretations for
``otherwise'' in (i) and (ii):
\begin{quote}
~(i) \parbox[t]{3.5in}{Q: How should I transport the dog?\\
    A:  You should {\sc carry} the dog. Otherwise you might get {\sc
    hurt}.}\\

\medskip

(ii) \parbox[t]{3.5in}{Q. What should I carry?\\
A. You should carry {\sc the dog}. Otherwise you might get {\sc hurt}.}\\
\end{quote}

\smallskip
\noindent
In both (i) and (ii), the questions constrain the theme/rheme
partition of the answer. Small capitals represent focus within the
rheme. In (i), the ``otherwise'' clause will be interpreted as warning the
hearer (H) that H might get hurt if s/he transports the dog in some
way other than carrying it (e.g., H might get tangled up in its
lead). In (ii),  the ``otherwise'' clause warns H that s/he might get
hurt if what she is carrying is not the dog (e.g., H might be walking
past fanatical members of the Royal Kennel Club).})

If we represent a conditional relation between two eventualities
with the asymmetric relation {\it if}($e_1$,$e_2$), where $e_1$
is derivved from the {\em antecedent} and $e_2$, from the
{\em consequent}, and we approximate a single contextually 
relevant alternative $e_2$ to an eventuality $e_1$ using a symmetric
{\em complement} relation, {\em complement}($e_1, e_2$) --
then we can represent the interpretation of ``otherwise'' as 
\begin{quote}
$\lambda$ x . {\em if}($VE$, x), where {\em complement}($VE,EV$)
\end{quote}
where variable $EV$ is resolved anaphorically to an eventuality in the
current discourse context that admits a complement. That is, ``otherwise''
requires a contextually relevant complement to its antecedent
and asserts that if that complement holds, the
argument to the $\lambda$-expression will. The resulting
$\lambda$-expression applies to the interpretation of the matrix
clause of ``otherwise'', resulting in the conditional being added to
the discourse context: 
\begin{quote}
$[\lambda$x . {\it if}($VE$,x)$]$ $\sigma$ $\equiv$
  {\em if}($VE$,$\sigma$), where {\em complement}($VE$,$EV$) 
\end{quote}
Here the relevant elements of (\ref{red1:ex}b) and
(\ref{red2:ex}b) can be represented as
\begin{quote}
$\alpha$ = otherwise\\
$R_\alpha$ = {\em if}\\
$S_{\ref{red1:ex}}$ = you get a ticket\\ 
$\sigma_{\ref{red1:ex}}$ = $e_3$, where $e_3$:get\_ticket(you)\\ 
$S_{\ref{red2:ex}}$ = go straight on\\
$\sigma_{\ref{red2:ex}}$ = $e_{3'}$, where $e_{3'}$:go\_straight(you)
\end{quote}
The unresolved interpretations of (\ref{red1:ex}b) and
(\ref{red2:ex}b) are thus:
\begin{quote}
$[\lambda$x . {\em if}($VE_{\ref{red1:ex}}$,x)$]\ e_3\ \equiv$
   {\em if}($VE_{\ref{red1:ex}}$,$e_3$), where {\em
   complement}($VE_{\ref{red1:ex}}$,$EV_{\ref{red1:ex}}$)\\ 
$[\lambda$x . {\em if}($VE_{\ref{red2:ex}}$,x)$]\ e_{3'}\ \equiv$
   {\em if}($VE_{\ref{red2:ex}}$,$e_{3'}$), where
   {\em complement}($VE_{\ref{red2:ex}}$,$EV_{\ref{red2:ex}}$)
\end{quote}
As we showed in Section~\ref{moresim:sec}, different ways of resolving
the anaphoric argument lead to different interpretations. In
(\ref{red1:ex}), the anaphoric argument is resolved to $e_2$:stop(you),
while in (\ref{red2:ex}), it is resolved to $e_{1}$:red(light1).
Thus the resulting interpretations of (\ref{red1:ex}b) and
(\ref{red2:ex}b) are, respectively
\begin{quote}
{\em if}($e_4$,$e_3$), where {\em complement}($e_2$,$e_4$) and
$e_2$:stop(you)\\ 
({\em If you do something other than stop, you'll get a ticket.})

{\em if}($e_{4'}$, $e_3$), where {\em complement}($e_1$,$e_{4'}$) and
$e_1$:red(light)\\
({\em If the light is not red, go straight on.})
\end{quote}

We have not been specific about how the anaphoric argument of
``otherwise'' (or of any other discourse adverbial) is resolved, other
than having it treated as a constraint satisfaction problem. This is
the subject of current and future work, 
exploring the empirical properties of resolution algorithms with data
drawn from appropriately annotated corpora and from psycholinguistic
studies of human discourse interpretation. To this end,
\namecite{creswell02} report on a preliminary annotation study
of discourse adverbials and the location and type of their
antecedents. This initial effort involves nine discourse adverbials --
three each from the classes of {\em concessive}, {\em
result} and {\em reinforcing} (additive) conjuncts given in
\cite{quirk72}. Meanwhile, \namecite{venditti02} present a preliminary
report on the use of a constraint-satisfaction model of
interpretation, crucially combining anaphoric and structural reasoning
about discourse relations, to predict subjects' on-line interpretation
of discourses involving stressed pronouns. 
In addition, two proposals have recently been
submitted to construct a larger and more extensively annotated corpus,
covering more adverbials, based on what we have learned from this
initial effort. This more extensive study would be an adequate basis
for developing
resolution algorithms.\footnote{With respect to how many discourse
adverbials there are, \namecite{quirk72} discuss 60 conjunctions and 
discourse adverbials under the overall heading ``time relations'' and
123 under the overall heading ``conjuncts''. The same entries appear
under several headings, so that the total number of conjunctions and
discourse adverbials they present is closer to 160. In another
enumeration of discourse adverbials, \namecite{forbes02:sigdial}
starts with 
all annotations of sentence-level adverbials in the Penn TreeBank, and
then filters them systematically to determine which draw part of their
meaning from the preceding discourse and how they do so. What we
understand from both these studies is that there are fewer than
200 adverbials to be considered, many of which are minor variations of
each other -- ``in contrast'', ``by contrast'', ``by way of
contrast'',  ``in comparison'', ``by comparison, ``by way of
comparison'' -- that are unlikely to differ in their anaphoric
properties, and some of which, such as ``contrariwise'', ``hitherto''
and ``to cap it all'', will occur only rarely in a corpus of modern
English.}

\subsection{Summary}

In this section, we have presented a general framework for anaphora
with the following features:
\begin{itemize}
\item Anaphors can access either one or more discourse referents
  or entities associated with them through bridging inferences. These
  are sufficient for interpreting anaphoric pronouns, 
  definite NPs and demonstrative NPs, allowing entities to be evoked
  by NPs or by clauses. In the case of clauses, this may be on an ``as
  needed'' basis, as in \cite{eckert01}.
\item A type of anaphor $\alpha$ that we call {\em
    lexically-specified} can also contribute additional meaning
  through a function $f_\alpha$ that is idiosyncratic to $\alpha$,
  that can be applied to either an existing discourse referent or an
  entity associated with it through a bridging inference. In the case
  of the premodifier ``other'', $f_\alpha$ applied to its argument
  produces contextually-relevant alternatives to that argument. In the
  case of the premodifier ``such'', it yields a set of entities that
  are similar to its argument in a contextually-relevant way.
\item Discourse adverbials are {\em lexically-specified} anaphors
  whose meaning function $f_\alpha$ is a $\lambda$-expression
  involving a binary relation $R_\alpha$ that is idiosyncratic to
  $\alpha$, one of whose arguments is resolved anaphorically and
  the other is provided compositionally, when the
  $\lambda$-expression is applied to $\alpha$'s matrix clause
  interpreted as an eventuality $\sigma$.
\end{itemize}
In the next section, we move on to consider how the presence of both a
semantic relation associated with a discourse adverbial and a semantic
relation associated with the adjacency of two clauses or a structural
connective between them, allows for interesting interactions between
the two.

\section{Patterns of Anaphoric Relations and Structural/Inferred Relations}
\label{inf:sec}

Prior to the current work, researchers have treated both explicit
structural connectives (coordinating and subordinating conjunctions,
and ``paired'' conjunctions) and discourse adverbials simply as
evidence for a particular structural relation
holding between adjacent units. For example, \namecite{kehler02}
takes ``but'' as evidence of a {\em contrast} relation between
adjacent units, ``in general'' as evidence of a {\em generalization}
relation, ``in other words'' as evidence of a {\em elaboration}
relation, ``therefore'' as evidence of a {\em result}
relation, ``because'' as evidence of a {\em explanation}
relation, and ``even though'' as evidence of a {\em denial of
  preventer} relation \cite[Chapter 2.1]{kehler02}.
Here Kehler has probably correctly identified the {\em type} of
relation that holds between elements, but not which elements it holds
between. 

In one respect, we follow previous researchers, in that we accept that
when clauses, sentences or larger discourse units are placed adjacent
to one another, listeners infer a relation between the two, and that
structural connective (coordinate or subordinate conjunction) gives
evidence for the relation that is intended to hold between them.

However, because we take discourse adverbials to contribute meaning
through an anaphoric connection with the previous discourse, this
means that there may be two relations on offer, and 
opens the possibility that the relation contributed by the discourse
adverbial can interact in more than one way with the relation conveyed
by a structural connective or inferred through adjacency. Below we
show that this prediction is correct. 

We start from the idea that -- in the absence of an explicit
structural connective -- defeasible inference correlates with
structural attachment of adjacent discourse segments in discourse
structure, relating their interpretations. The most basic relation is
that the following segment in some way describes the same
object or eventuality as the one it abuts ({\em elaboration}). But
evidence in the segments can lead (via defeasible inference) to a more
specific relation, such as one of the {\em resemblence} relations
(e.g., {\em parallel}, {\em contrast}, 
{\em exemplification}, {\em generalisation}), or {\em cause-effect}
relations ({\em result}, {\em explanation}, {\em violated
expectation}), or {\em contiguity} relations ({\em narration})
described in  \cite{hobbs90,kehler02}.
If nothing more specific can be inferred, the relation will remain simply
{\em elaboration}. What explicit structural connectives can do is
convey relations that are not easy to convey by defeasible inference (e.g.,
``if'', conveying {\em condition}, and ``or'', conveying {\em
  disjunction}) or provide non-defeasible evidence for
an inferrable relation (e.g., ``yet'', ``so'' and ``because'').

Discourse adverbials can {\em interact} with structural connectives,
with adjacency-triggered defeasible inference and with each other. To
describe the ways in which we have so far observed discourse
adverbials to interact with relations conveyed structurally, we extend
the notation used in the previous section:  
\begin{itemize}
\item $\alpha$ = discourse adverbial;
\item $R_{\alpha}$ = the name of the relation associated with $\alpha$.
\item $S$ = the matrix clause/sentence of $\alpha$;
\item $\sigma$ = the logical form (LF) interpretation of $S$;
\end{itemize}
adding the following:
\begin{itemize}
\item $D$ = the discourse unit that is left-adjacent to $S$, to which
  a relationship holds either by inference or a structural connective;
\item $\delta$ = the LF interpretation of $D$;
\item $R$ = the name of the relation that holds with $\delta$;
\end{itemize}
While $\delta$ is one argument of $R$, we show below that its other
argument may be one of at least two different abstract objects.

\subsubsection*{Case 1: $\sigma$ separately serves as an argument to
  both $R_{\alpha}$ and $R$.}

This is the case that holds in Example~\ref{barolo-rep:ex} (repeated
below).
\begin{quote}
(\ref{barolo-rep:ex})\ \ \parbox[t]{4.5in}{a. John loves Barolo.\\
b. So he ordered three cases of the '97.\\
c. But he had to cancel the order\\
d.  because he {\em then} discovered he was broke.}\\
\end{quote}
We have already seen that the interpretation of the clause in
(\ref{barolo-rep:ex}d) following ``because'' involves: 
\begin{quote}
$R_{\alpha}$ = {\em after}\\ 
$\sigma$ = $e_4$:discover(j,$e_5$), where $e_5$:broke(j)\\
$[\lambda$x . {\em after}(x,$EV$)$] e_4$ $\equiv$  {\em after}($e_4,EV$)
\end{quote}
where $EV$ is resolved to $e_2$:order(j, $c_1$), and the
proposition {\em after}($e_4,e_2$) is added to the discourse context
-- i.e., John's discovering he was broke is after his ordering the wine.

Now consider the {\em explanation} relation $R$ associated with
``because'' in (\ref{barolo-rep:ex}d). It relates  $e_4$, John's
finding he was broke, to the intepretation of (\ref{barolo-rep:ex}c),
$e_3$:cancel(j,$o_1$) -- that is, {\em explanation}($e_4$,$e_3$).
Clause~\ref{barolo-rep:ex}d thus adds both {\em
  explanation}($e_4$,$e_3$) and
{\em after}($e_4,e_2$) to the discourse. While these two propositions
share an argument ($e_4$), they are nevertheless
distinct.\footnote{Because eventuality $e_4$ ``John's finding he was
  broke'' both {\em explains} the cancelling and {\em follows} the
  ordering, it follows that the cancelling is after the ordering.}

\subsubsection*{Case 2:  $R_{\alpha}(\sigma,e_i)$ is an argument of
  $R$.}

In Case 1, it is the interpretation of the adverbial's matrix clause
$\sigma$ that serves as one argument to the discourse relation $R$. In
contrast, in Case 2, that argument is filled by the
{\em relation} contributed by the discourse adverbial (itself an
abstract object available for subsequent reference). In both cases,
the other argument to $R$ is $\delta$. 

One configuration in which Case 2 holds is with the discourse
adverbial ``otherwise''. Recall from Section~\ref{semcontrib:sec} that
the interpretation of ``otherwise'' involves a conditional relation
between the complement of its anaphoric argument and the
interpretation $\sigma$ of its matrix clause: 
\begin{quote}
$[\lambda$x . {\it if}($VE$,x)$]$ $\sigma$ $\equiv$
  {\em if}($VE$,$\sigma$), where complement($VE$,$EV$)
\end{quote}
With variable $EV$ resolved to an eventuality in the discourse
context, it is the resulting relation (viewed as an abstract object)
that serves as one argument to 
$R$, with $\delta$ serving as the other. We can see this most clearly
by considering variants of examples (\ref{red1:ex}) 
and (\ref{red2:ex}) that contain an explicit connective between the
clauses. In (\ref{red1:ex}), the conjunction ``because'' is made
explicit (Example~\ref{red1-rep:ex}), while in (\ref{red2:ex}), the
connective is simply ``and'' or ``but'' (Example~\ref{red2-rep:ex}).
\enumsentence{\label{red1-rep:ex}
If the light is red, stop, because {\em otherwise} you'll get a
  ticket.\\
$R_\alpha$ = {\em if}\\
$\sigma_{\ref{red1:ex}}$ = $e_3$:get\_ticket(you)}
\enumsentence{\label{red2-rep:ex}
If the light is red, stop, and/but {\em otherwise} go straight on.\\
$R_\alpha$ = {\em if}\\
$\sigma_{\ref{red2:ex}}$ = $e_{3'}$:go\_straight(you)}

\noindent
In the case of ({\ref{red1-rep:ex}), resolving ``otherwise''
contributes the relation
\begin{quote}
$e_6$: {\em if}($e_4$,$e_3$), where {\em complement}($e_4$,$e_2$) and
$e_2$:stop(you)\\ 
({\em If you do something other than stop, you'll get a ticket.}) 
\end{quote}
At the level of logical form (LF), the abstract object $e_6$ that is
associated with the conditional relation serves as 
one argument to the {\em explanation} relation contributed by
``because'', with $e_2$ being the other. That is, ``because'' and
``otherwise'' together end up contributing {\em explanation}($e_2$,$e_6$)
(i.e., your needing to stop is explained by the fact that if you do
something other than stop, you'll get a ticket).

In the case of ({\ref{red2-rep:ex}), resolving ``otherwise''
contributes the relation 
\begin{quote}
$e_{6'}$:{\em if}($e_{4'}$, $e_{3'}$), where {\em
  complement}($e_{4'}$,$e_1$) and $e_1$:red(light)\\
({\em If the light is not red, go straight on.})
\end{quote}
What is the discourse relation to which ``otherwise'' contributes this
abstract object $e_{6'}$?  Whether the connective is ``and'' or
``but'', both its conjuncts describe (elaborate) alternative
specializations of the same situation $e_0$ introduced earlier in the
discourse (e.g., $e_0$ could be associated with the first sentence of
``Go another mile and you'll get to a bridge. If the
light is red, stop. Otherwise go straight on.'') If the connective is
``and'', what is added to context might simply be {\em
  elaboration}($e_{6'}$,$e_0$). (N.B. Without ``otherwise'', the
relation {\em elaboration}($e_5$,$e_0$) would have been added to
context, where $e_5$ is the abstract object associated with the
interpretation of
``If the light is red, stop''.) If the connective is ``but'', then
one might also possibly add {\em contrast}($e_{6'}$,$e_{5}$) --
i.e., The situation that $[$if the light is red$]$ you should stop is
in contrast with the situation that if the light is not red, you
should go straight on.\footnote{A much finer-grained treatment of
the semantics of ``otherwise'' in terms of context-update potential is
given in \cite{kkw01}. Here we are just concerned with its interaction
with structural connectives and adjacency-triggered relations.}

As is clear from the original pair of examples (\ref{red1:ex}) and
(\ref{red2:ex}), similar interpretations can arise through
adjacency-triggered inference as arise with an explicit connective.
In either case, the above treatment demonstrates that there is no need
for a separate {\em otherwise} relation, as proposed in Rhetorical
Structure Theory \cite{mt88}. We are not, however, entirely clear at
this point when Case 1 holds and when Case 2 does. A more careful
analysis is clearly required.

\subsubsection*{Case 3:  $R_{\alpha}$ is parasitic on $R$.} 

{\bf Case 3} appears to hold with discourse adverbials such as ``for
example'' and ``for instance''. Their interpretation appears to be
{\em parasitic} on the relation associated with a structural
connective or discourse adverbial to their left, or on an 
inferred relation triggered by adjacency. The way to understand this
is to first consider {\em intra-clausal} ``for example'', where it
follows the verb, as in  
\enumsentence{Q. What does this box contain?\\
A. It contains, {\em for example}, some hematite. \label{iron:ex}}
The interpretation of ``for example'' here involves abstracting the
meaning of its matrix structure with respect to the material to its
right, and then making an assertion with respect to this
abstraction. That is, if the LF contributed by the
matrix clause of (\ref{iron:ex}A) is, roughly, 
\begin{quote}
{\bf i.} contain(box1,hematite1)
\end{quote}
then the LF resulting from the addition of ``for example'' can be
written either with set notation (as in {\bf ii}), taking an entity to
exemplify a set, or with $\lambda$-notation (as in {\bf iii}), taking
an entity to exemplify a property:
\begin{quote}
{\bf ~ii.} {\em exemplify}(hematite1, \{X $\mid$ contain(box1,X)\})\\
{\bf iii.} {\em exemplify}(hematite1, $\lambda$X . contain(box1,X))
\end{quote}
Both express the fact that ``hematite'' is an example of what is
contained in the box.\footnote{The material to the right of ``for
example'' can be any kind of constituent, including such strange
ones as
\begin{quote}
John gave, for example, a flower to a nurse.
\end{quote}
Here, ``a flower to a nurse'' would be an example of the set of
object-recipient pairs within John's givings. Such non-standard
constituents are also found with coordination, which was one motivation
for Combinatorial Categorial Grammar  \cite{steed96}. This just
illustrates another case where such non-standard constituents are
needed.} Since one can derive ({\bf i}) logically from either ({\bf
ii}) or ({\bf iii}), one might choose to retain only ({\bf ii}) or
({\bf iii}) and derive ({\bf i}) if and when it is needed. In the
remainder of the paper, we use the $\lambda$ notation given in
({\bf iii}). 
Notice that
from the 
perspective of compositional semantics, ``for example'' resembles a
quantifier, in that the scope of its interpretation is not isomorphic
to its syntactic position. Thus producing an interpretation for ``for
example'' will require similar techniques to those used in
interpreting quantifiers. We will take this up again in
Section~\ref{tag:sec}. 

If we look at the comparable situation in discourse such as
(\ref{sofor:ex})--(\ref{trustbecause:ex}), where ``for example''
occurs to the right of a discourse connective,
it can also be seen as abstracting the interpretation of its
discourse-level matrix structure, with respect to the material to its
right.
\enumsentence{
John just broke his arm. {\em So, for example}, he can't
cycle to work now. \label{sofor:ex}}
\enumsentence{You shouldn't trust John {\em because, for example}, he
never returns what he borrows. \label{trustbecause:ex}}
In (\ref{sofor:ex}), the connective  ``so'' leads to
\begin{quote}
{\em result}($\sigma$,$\delta$)
\end{quote}
being added to the discourse, where $\sigma$ is the interpretation of
``John can't cycle to work now'', and $\delta$ is the interpretation
of ``John just broke his arm''. ``For example'' then abstracts this
relation with respect to the material to its right (i.e., $\sigma$),
thereby contributing:
\begin{quote}
{\em exemplify}($\sigma$, $\lambda$X . {\em result}(X, $\delta$))
\end{quote}
That is, ``John can't cycle to work'' is an example of what results
from ``John breaking his arm''.
Similarly, ``because'' in (\ref{trustbecause:ex}) leads to
\begin{quote}
{\em explanation}($\sigma$,$\delta$)
\end{quote}
being added to the discourse, where $\sigma$ is the interpretation of
``he never returns what he borrows'',  $\delta$ is the interpretation of
``you shouldn't trust John'', and ``for example'' adds
\begin{quote}
{\em exemplify}($\sigma$, $\lambda$X . {\em explanation}(X,$\delta$))
\end{quote}
i.e., that $\sigma$ is an example of the reasons for not trusting John.

``For example'' interacts with discourse adverbials in the same way:
\enumsentence{Shall we go to the Lincoln Memorial? {\em Then, for
    example}, we can go to the White House.\label{whitehouse:ex}}
\enumsentence{As a money manager and a grass-roots environmentalist, I
  was very disappointed to read in the premiere issue of Garbage that
  The Wall Street Journal uses 220,000 metric tons of newsprint each
  year, but that only 1.4\% of it comes from recycled paper. 
  {\em By contrast}, the Los Angeles Times, {\em for example}, uses
  83\% recycled paper. $[$WSJ, from Penn TreeBank /02/wsj-0269$]$
  \label{paper:ex}}
In Example~\ref{whitehouse:ex}, the resolved discourse adverbial
``then'' leads to {\em after}($\sigma$,$\delta$)
being added to the discourse context, where $\sigma$ is the interpretation of
``we can go to the White House'', $\delta$ is the interpretation of
``we can go to the Lincoln Memorial'', and ``for example'' adds
\begin{quote}
{\em exemplify}($\sigma$, $\lambda$X . {\em after}(X,$\delta$))
\end{quote}
i.e., that $\sigma$ is an example of the events that $[$can$]$ follow
going to the Lincoln Memorial. 
({\bf N.B.} As already noted, we are being fairly fast and loose
regarding tense and modality, in the interests of focussing on the
types of interactions.)

In Example~\ref{paper:ex}, the resolved discourse anaphor ``by
contrast'' contributes {\em contrast}($\sigma$,$\delta$), where
$\sigma$ is the interpretation of ``the LA Times using 83\% recycled
paper'' and $\delta$ is the intepretation of ``only 1.4\% of it
$[$newsprint used by the WSJ$]$ comes from recycle paper''. ``For
example'' then contributes
\begin{quote}
{\em exemplify}($\sigma$, $\lambda$X . {\em contrast}(X,$\delta$))
\end{quote}
i.e., that $\sigma$ is one example of contrasts with the
WSJ's minimal use of recycled paper.

What occurs with discourse connectives and adverbials can also occur
with relations added through adjacency-triggered defeasible inference,
as in 
\enumsentence{You shouldn't trust John. {\em For example}, he never
  returns what he borrows.\label{trustrep:ex}\\
{\em explanation}($\delta$,$\sigma$)\\
{\em exemplify}($\sigma$, $\lambda$X . {\em explanation}($\delta$,X))
}
Here, as in (\ref{trustbecause:ex}), the relation provided by
adjacency-triggered inference is $R$={\em explanation}, which is then
used by ``for example''.

But what about the many cases where only {\em exemplify} seems
present, as in  
\enumsentence{In some respects they $[$hypertext books$]$ are clearly superior
  to normal books, for example they have database cross-referencing
  facilities ordinary volumes lack. $[${\bf British National Corpus},
  CBX 1087$]$ \label{forex1:ex}}
\enumsentence{He $[$James Bellows$]$ and his successor, Mary Anne
  Dolan, restored respect for the editorial product, and though in
  recent years the paper had been limping along on limited resources,
  its accomplishments were notable. {\em For example}, the Herald
  consistently beat its much-larger rival on disclosures about Los
  Angeles Mayor Tom Bradley's financial dealings. \label{forex2:ex}}

There are at least two explanations: One is that
``for example'' simply provides direct non-defeasible evidence for
{\em exemplify}, which is the only relation that holds. The other
explanation follows the same pattern as the examples given above, but
with no further relation than {\em elaboration}($\sigma$,$\delta$).
That is, we 
understand in (\ref{forex1:ex}) that ``having database cross-referencing
facilities'' {\em elaborates} the respects in which hypertext books are
superior to normal books, while in (\ref{forex2:ex}), we understand 
that ``the Herald $[$newspaper$]$ consistently beating its much-larger
rival'' {\em elaborates} the claim that ``its accomplements were notable''.
This {\em elaboration} relation is then abstracted 
(in response to ``for example'') to produce:
\begin{quote}
{\em exemplify}($\sigma$, $\lambda$X . {\em elaboration}(X, $\delta$))
\end{quote}
i.e., that this is one example of many possible elaborations.
Because this is more specific than {\em elaboration} and seems to mean
the same as {\em exemplify}($\sigma$,$\delta$), one might simply take
it to be the only relation that holds. Given that
so many naturally-occuring instances of ``for example'' occur with
{\em elaboration}, it is probably 
useful to persist with the above shorthand. But it shouldn't obscure
the regular pattern that appears to hold.

Before going on to {\bf Case 4}, we should comment on an ambiguity
associated with ``for example''. When ``for example'' occurs after an
NP, PP or clause that can be interpreted as a general concept or a
set, it can contribute a relation between the general concept/set
and an instance, rather than being parasitic on another relation. For
example, in:
\enumsentence{In the case of the managed funds they will be denominated in
a leading currency, for example US dollar, \ldots $[${\bf BNC} CBX 1590$]$
\label{dollar:ex}}
``for example''  relates the general concept denoted by ``a leading
currency'' to a specific instance, US dollars.
(In ``British'' English, the BNC shows that most such examples occur
with ``such as'' -- i.e., 
in the construction ``such as for example''. This paraphrase does not
  work with the predicate-abstracting ``for example'' that is of
  primary concern here, such as in Example~\ref{iron:ex}.)

But ``for example'' occurring after an NP, PP or clause can,
alternatively, contribute
a more subtle parasitic relationship to the previous clause, as in
\enumsentence{All the children are ill, so Andrew, {\em for example},
  can't help out in the shop. \label{subtle1:ex}}
This differs from both  (\ref{sofor:ex}) and (\ref{dollar:ex}).
That is, one cannot paraphrase (\ref{subtle1:ex}) as
(\ref{subtle2:ex}) as in (\ref{sofor:ex}) where ``for example''
follows ``so'':
\enumsentence{All the children are ill, so {\em for example}
  Andrew can't help out in the shop.\label{subtle2:ex}}
(\ref{subtle2:ex}) simply specifies an example
consequence of all the children being ill, as does
\enumsentence{All the children are ill, so {\em for example} one of us
  has to be at home at times.}
In contrast, (\ref{subtle1:ex}) specifies an example consequence for
Andrew, {\em as one of the children}. Support for this comes from the
fact that in (\ref{subtle2:ex}), Andrew doesn't have to be one of the
children: he could be their nanny or child minder, now stuck with
dealing with alot of sick kids. But (\ref{subtle1:ex}) is not
felicitous if Andrew is not one of the children.

We suspect here the involvement of {\em Information Structure}
\cite{steed01:LI}: While the interpretation conveyed by ``for
example'' is parasitic on the adjacency relation ({\em result} in
Example~\ref{subtle1:ex}), its position after the NP ``Andrew'' in
(\ref{subtle1:ex}) may indicate a {\em contrastive theme} with respect
to the previous clause, according to which Andrew {\em in contrast to
  the other children} suffers this particular consequence. But more
work needs to be done on this to gain a full understanding of what is
going on.

\subsubsection*{Case 4: $R_{\alpha}$ is a defeasible rule that
  incorporates $R$.} 

Case 4 occurs with discourse adverbials that carry the same
presupposition as the discourse connectives ``although'' and the
concessive sense of ``while'' \cite{lagerwerf98}. Case 4 shares one
feature with Case 1, in that the discourse relation $R$
conveyed by a structural connective or inferred from adjacency holds
between  $\sigma$ (the interpretation of the adverbial's
matrix clause) and $\delta$ (the interpretation of the left-adjacent
discourse unit). Where it differs is that the result is then
incorporated into the {\em  presupposition} of the discourse
adverbial. This presupposition, according to \namecite{lagerwerf98},
has the nature of a presupposed (or conventionally implicated)
{\em defeasible rule} that fails to hold in the current situation. He
gives as an example
\enumsentence{Although Greta Garbo was called the yardstick of beauty,
  she never married.\label{garbo:ex}}
This asserts both that Greta Garbo was called the yardstick of beauty
and that she never married. The first implies that Greta Garbo is
beautiful. The example also presupposes that, in general, 
if a woman is beautiful, she will marry. If such
a presupposition can be accommodated, it will simply be added to the
discourse context. If not, the hearer will find the utterance
confusing or possibly even insulting.

We argue here that the same thing happens with the discourse
adverbials ``nevertheless'' and ``though''. The difference
is that, with discourse adverbials, 
the {\em antecedent} to the rule derives anaphorically from the
previous discourse, while the {\em consequent} derives from the
adverbial's matrix clause. (With the conjunctions ``although'' and
concessive ``while'', both arguments are provided structurally.)

Here we first illustrate Case 4 with two examples in which
``nevertheless'' occurs in the main clause of a sentence containing a
preposed subordinate clause. The subordinate conjunction helps clarify
the relation between the clauses that forms the basis for the
presupposed defeasible rule. After this, we give a further example
where the relation between the adjacent clauses comes through
inference.
\enumsentence{{\em While} John is discussing politics, he
  is {\em nevertheless} thinking about his fish. \label{chess:ex}}
In ({\ref{chess:ex}), the conjunction ``while'' conveys a temporal
  relation $R$ between the two clauses it connects
\begin{quote}
{\em during}($e_2,e_1$), where $e_1$:discuss(john,politics) and
$e_2$:think\_about(john,fish)
\end{quote}
What ``nevertheless'' contributes to (\ref{chess:ex}) is a defeasible
rule based on this relation, which we will write informally as
\begin{quote}
 {\em during}($X$,$E$) $\wedge$ $E$:discuss($Y$,politics)) $>$
 $\neg X$:think\_about($Y$,fish))\\
{\em Normally, whatever one does during the time one is discussing
 politics, it is not thinking about one's fish.}
\end{quote}
This rule uses Asher and Morreau's \shortcite{ashmorr91} defeasible
implication operator 
($>$) and abstracts over the individual (John), which seems
appropriate for the general statement conveyed by the present tense
of the utterance.

Similarly, in
\enumsentence{Even {\em after} John has had three glasses of wine, he is
{\em nevertheless} able to solve difficult math
problems. \label{drink:ex}}
the conjunction ``after'' contributes a relation between the two
clauses it connects
\begin{quote}
{\em after}($e_2, e_1$), where $e_1$:drink(john,wine) and
$e_2$:solve(john,hard\_problems)
\end{quote}
What ``nevertheless'' contributes to this example is a defeasible rule
that we will again write informally as
\begin{quote}
{\em after}($X$,$E$) $\wedge$ $E$:drink($Y$,wine)) $>$
$\neg X$:solve($Y$,hard\_problems))\\ 
{\em Normally, whatever one is able to do after one has had three
glasses of wine, it is not solving difficult algebra problems.}
\end{quote}
Again, we have abstracted over the individual, as the presupposed
defeasible rule associated with the present tense sentence appears to
be more general than a statement about a particular
individual.\footnote{We speculate that the reason such examples such
  as (\ref{chess:ex}) and (\ref{drink:ex}) sound more natural with 
the focus particle ``even'' applied to the subordinate clause, is that
``even'' conveys an even greater likelihood that the defeasible rules
holds, so ``nevertheless'' emphasises its failure to do so.}

On the other hand, in the following example illustrating a presupposed
defeasible rule and a discourse relation associated with adjacency, it
seems possible for the presupposed defeasible rule to be about John
himself.
\enumsentence{John is discussing politics. {\em Nevertheless}, he
  is thinking about his fish.\label{chess2:ex}}
Here the discourse relation between the two clauses, each of which
denotes a specific event, is
\begin{quote}
{\em during}($e_2,e_1$), where $e_1$:discuss(john,politics) and
$e_2$:think\_about(john,fish)
\end{quote}
(N.B. Our LF representation isn't sufficiently rich to express the
difference between (\ref{chess:ex}) and (\ref{chess2:ex}).)
What ``nevertheless'' contributes here is the presupposed defeasible
rule
\begin{quote}
{\em during}($X$,$e_1$) $>\  \neg X$ = $e_2$\\
{\em Normally what occurs during John's discussing politics is not
  John thinking about his fish.}
\end{quote}

\namecite{lagerwerf98} does not discuss how specific or general will
be the presupposed defeasible rule that is accommodated, nor what
factors affect the choice. \namecite{kw01:iwcs} also punt on the
question, when considering the effect of Information Structure on what
presupposed defeasible rule is associated with ``although''. Again,
this seems to be a topic for future work.

\subsection*{Summary}

We have indicated four ways in which we have found the relation
associated with a discourse adverbial to interact with a relation $R$
triggered by adjacency or conveyed by structural connectives or, in
some cases, by another relational anaphor:
\begin{enumerate}
\item  $\sigma$ separately serves as an argument to both $R_{\alpha}$
  and $R$;
\item  $R_{\alpha}(\sigma,e_i)$ is an argument of $R$;
\item  $R_{\alpha}$ is parasitic on $R$;
\item  $R_{\alpha}$ is a defeasible rule that incorporates $R$. 
\end{enumerate}

We do not know whether this list is exhaustive or whether a discourse
adverbial always behaves the same way vis-a-vis other relations.
Moreover, in the process of setting down the four cases we discuss, we
have identified several problems that we have not addressed, on which
further work is needed. Still, we hope that we have convinced the
reader of our main thesis -- that by recognizing discourse adverbials
as doing something different from simply signalling the discourse
relation between adjacent discourse units and by considering their
contribution as relations in their own right, one can begin to
characterise different ways in which anaphoric and structural
relations may themselves interact.

\section{Lexicalised Grammar for Discourse Syntax and Semantics}
\label{tag:sec}

The question we consider in this section is how the treatment we
have presented of discourse adverbials and structural connectives
can be incorporated into a general approach to discourse
interpretation. There are three possible ways.

The first possible way is to simply incorporate our treatment of
adverbials and connectives into a sentence-level grammar, since such
grammars already cover the syntax of sentence-level conjunction (both
coordinate and subordinate) and the syntax of adverbials of all types.
The problem with this is that sentence-level grammars -- whether phrasal
or lexicalized -- stop at explicit sentence-level conjunction and do not
provide any mechanism for forming the meaning of multi-clausal
units that cross sentence-level punctuation. Moreover, as we have already
seen in Section~3, the interpretation of discourse adverbials can
interact with the implicit relation between adjacent sentences, as
well as with an explicitly signalled relation, so that
a syntax and compositional semantics that stops at the sentence will
not provide all the structures and associated 
semantics needed to build the structures and interpretations of interest.

The second possibility is to have a completely different approach
to discourse-level syntax and semantics than to sentence-level
syntax and semantics, combining (for example) a Definite Clause
Grammar with Rhetorical Structure Theory. But as we and others have
already noted, this requires discourse semantics reaching further
and further into sentence-level syntax and semantics to handle
relations between main and embedded clauses, and between embedded
clauses themselves, as in Example~\ref{drunk:ex}.
\enumsentence{
If they're drunk and they're meant to be on parade and you go to
  their room and they're lying in a pool of piss, then you lock them
  up for a day.\\ $[$The Independent, 17 June 1997$]$
\label{drunk:ex}}
Thus it becomes harder and harder to distinguish the scope of
discourse-level syntax and semantics from that at the sentence-level.

The third possibility is to recognize the overlapping scope and
similar mechanisms and simply extend a sentence-level grammar and its
associated semantic mechanisms to discourse. Its additional
responsibilities would be to account for the formation
of larger units of discourse from smaller units; the projection of
discourse unit interpretation onto the interpretation of the larger
discourse units they participate in; and the effect of discourse
unit interpretation on the evolving discourse model.
There are two styles of grammar one could use for this -- (a) a
phrase-structure grammar (PSG), which is what \namecite{pol-van96} use
for discourse, or (b) a lexicalized grammar that extends to discourse,
a sentence-level lexicalized grammar such as  Tree-Adjoining Grammar
\cite{joshi87,tech-rept01} or Combinatory Categorial Grammar (CCG)
\cite{steed96,steed00}. 

\begin{figure}
\begin{quote}
\textsf{Seg} := \parbox[t]{3.5in}{\textsf{SPunct} \textsf{Seg} $\mid$
       \textsf{Seg} \textsf{SPunct} $\mid$ \textsf{SPunct} $\mid$ \\
           {\em on the one hand} \textsf{Seg} {\em on the other hand} \textsf{Seg} $\mid$\\
           {\em not only} \textsf{Seg} {\em but also} \textsf{Seg}}

\smallskip

\textsf{SPunct} := \textsf{S Punctuation}

\textsf{Punctuation} := . $\mid$ ; $\mid$ : $\mid$ ? $\mid$ !

\textsf{S} := \parbox[t]{3.5in}{\textsf{S} \textsf{Coord} \textsf{S} $\mid$
     \textsf{S} \textsf{Subord} \textsf{S} $\mid$
     \textsf{Subord} \textsf{S} \textsf{S} $\mid$ \textsf{Sadv} \textsf{S} $\mid$\\
      NP \textsf{Sadv} VP $\mid$ \textsf{S} \textsf{Sadv}}

\smallskip

\textsf{Coord} := \emph{and} $\mid$ \emph{or} $\mid$ \emph{but} $\mid$ \emph{so}

\textsf{Subord} := \emph{although} $\mid$ \emph{after} $\mid$ \emph{because}
    $\mid$ \emph{before} $\mid$ ...

\textsf{Sadv} := \textsf{DAdv} $\mid$ \textsf{SimpleAdv}

\textsf{DAdv} := \emph{instead} $\mid$ \emph{otherwise} $\mid$ \emph{for
   example} $\mid$ \emph{meanwhile} $\mid$ ...

\textsf{SimpleAdv} := \emph{yesterday} $\mid$ \emph{today} $\mid$
    \emph{surprisingly}  $\mid$ \emph{hopefully} $\mid$ ... 
\end{quote}
\caption{PS rules for a discourse grammar}
\label{gram:fig}
\end{figure}

The latter is what we argue for, even though TAG and CCG are weakly
context-sensitive (CS) and the power needed for a discourse grammar
with no crossing dependencies  is only CF
(Section~\ref{no-cross:sec}). Our argument is based on our desire to
use a discourse grammar in Natural Language
Generation (NLG). It is well-known that context-free PSGs (CF PSGs)
set up a complex search space for NLG. A discourse 
grammar specified in terms of phrase structure rules such as those
shown in Figure~\ref{gram:fig} doesn't provide sufficient guidance
when reversed to use in generating discourse. For example,  one might
end up having to guess randomly how many 
sentences and connectives one had, in what order, before being able to
fill the sentences and connectives in with any content.
More generally, trying to generate exactly a given semantics when semantics
underspecifies syntactic dependency (as discourse semantics must, on our
account) is known to be intractable \cite{kollstr02}.  An effective
solution is to generate semantics and syntax simultaneously, which is
straightforward with a lexicalized grammar \cite{sdwbp01}.

Given the importance of various types of inference in discourse
understanding, there is a second argument for using a lexicalized
discourse grammar, which
derives from the role of implicature in discourse. Gricean reasoning
about implicatures requires a hearer be able to infer
the meaningful alternatives that a speaker had in composing a sentence.
With lexicalization, these alternatives can be given by a grammar,
allowing the hearer, for example, to ask sensible questions like
``Why did the speaker say 'instead' here instead of nothing at all?''
and draw implicatures from this. A CF PSG, on the other hand,
might suggest questions like ``Why did the
speaker say two sentences rather than one here?'', which seem empirically
not to lead to any real implicatures. (On the contrast between choices,
which seem to lead to implicatures, and mere alternative linguistic
formulations, which do not seem to, see for example
\cite{dr95,levinson00}.)

In several previous papers \cite{wkj99b,wksj99,wksj99:lorid},
we described how our approach fits into the framework of
Tree Adjoining Gramar. This has led to the initial version of a
discourse parser \cite{dltag01} in which the same parser that
builds trees for individual clauses using clause-level LTAG trees,
then combines them using discourse-level LTAG trees. Here we simply
outline the grammar, called DLTAG (Section~\ref{tagsyn:sec}),
and then show how it supports the approach to structural and anaphoric
discourse connectives presented earlier (Section~\ref{tagex:sec}).

(Of course, one still needs to account for how speakers realise their
intentions through text and how what is achieved through a single unit
of text contributes to what a speaker hopes to achieve through any
larger unit it is embedded in. Preliminary accounts are given in
\cite{gs90,mm96}. However, given the complex relation between
individual sentences and speaker intentions, it is unlikely that the
relation between multi-sentence discourse and speaker intentions can
be modelled in a straightforward way similar to the
basically monotonic compositional process that we have discussed
in this paper for discourse semantics.)

\subsection{DLTAG and Discourse Syntax}
\label{tagsyn:sec}

A lexicalized TAG begins with the notion of a lexical {\em anchor},
which can have one or more associated tree structures. For example,
the verb {\em likes} anchors one tree corresponding to
{\em John likes apples}, another corresponding to the topicalized
{\em Apples John likes}, a third corresponding to the passive
{\em Apples are liked by John}, and others as well. That is,
there is a tree for each minimal syntactic construction in which
{\em likes} can appear, all sharing the same predicate-argument structure.
This syntactic/semantic encapsulation is possible because of the extended
domain of locality of LTAG.

A lexicalized TAG contains two kinds of {\em elementary} trees:
{\em initial} trees that reflect basic functor-argument dependencies
and {\em auxiliary} trees that introduce recursion and allow
elementary trees to be modified and/or elaborated.
Unlike the wide variety of trees needed at the clause level,
we have found that extending a lexicalized TAG to discourse
only requires a few elementary tree structures, possibly because
clause-level syntax exploits structural variation in ways that
discourse doesn't.

\subsubsection{Initial Trees}

\begin{figure*}
\centerline{\epsfig{file=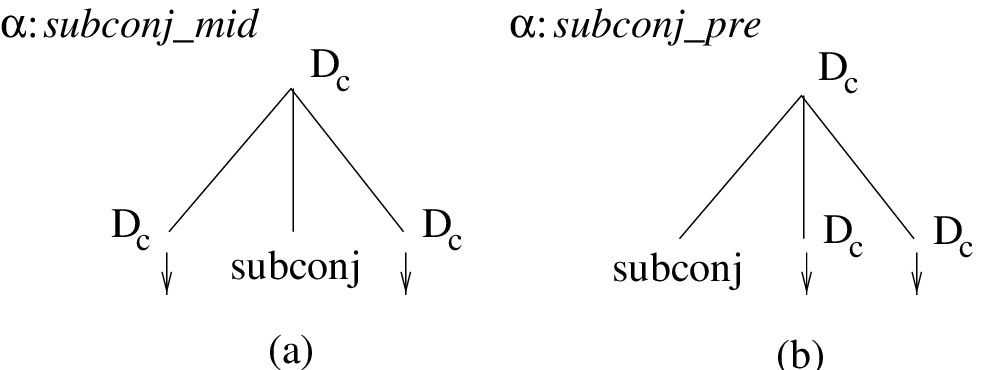, scale=0.75}}
\caption{Initial trees (a-b) for a subordinate conjunction.
{\bf D}$_c$ stands for ``discourse clause'', $\downarrow$ indicates a
substitution site, while ``subconj'' stands for
the particular subordinate conjunction that anchors the tree.}
\label{subord:fig}
\end{figure*}

DLTAG has initial trees associated with
subordinate conjunctions, with parallel constructions,
and with some coordinate conjuctions. We describe each in turn. 

In the large LTAG developed by the XTAG project \cite{tech-rept01},
subordinate clauses are seen as {\em adjuncts} to sentences or
verb phrases -- i.e., as auxiliary trees -- because they are outside
the domain of locality of the verb. In DLTAG, however, it is
predicates on clausal arguments (such as 
coordinate and subordinate conjunctions) that define the domain of
locality. Thus, at this level, these predicates anchor {\em initial
trees} into which clauses substitute as
arguments. Figure~\ref{subord:fig} shows the initial trees
for postposed subordinate clauses (a) and preposed subordinate
clauses (b).\footnote{While in an earlier paper \cite{wj98}, we
discuss reasons for taking the lexical anchors of the initial
trees in Figures~\ref{subord:fig} and \ref{par:fig} to be feature
structures, following the analysis in 
\cite{knott96,knott-mellish96}, here we just take them to be
specific lexical items.}
At both leaves and root is a {\em discourse clause} (D$_c$) -- a clause
or a structure composed of discourse clauses.

\begin{figure*}
\centerline{\epsfig{file=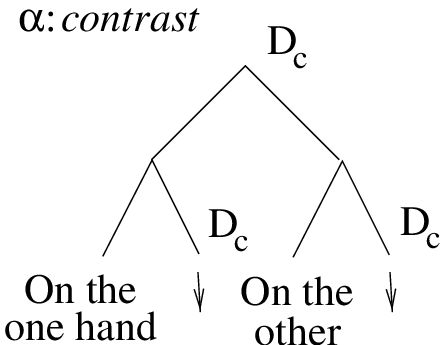,scale=.75}}
\caption{An initial tree for parallel constructions. This particular
one is for a contrastive construction anchored by ``on the one hand''
and ``on the other hand''.}
\label{par:fig}
\end{figure*}

One reason for taking something to be an initial tree is that its
local dependencies can be stretched long-distance. At the sentence-level,
the dependency between {\em apples} and {\em likes} in
{\em apples John likes} is localized in all the trees for {\em likes}.
This dependency can be stretched long-distance, as in
{\em Apples, Bill thinks John may like}. In discourse, as we noted
in Section~\ref{anaph:sec}, local dependencies can be stretched
long-distance as well -- as in
\eenumsentence{
\item Although John is generous, he's hard to
       find.
\item Although John is generous -- for example, he gives money to
  anyone who asks him for it -- he's hard to find.
}
\eenumsentence{ \label{par:ex}
\item On the one hand, John is generous. On the other hand, he's
    hard to find.
\item On the one hand, John is generous. For example, suppose you
  needed some money: You'd only have to ask him for it. On the other
  hand, he's hard to find.}
Thus DLTAG also contains initial trees for parallel constructions as
in (\ref{par:ex}). Such an initial tree is shown in Figure~\ref{par:fig}. 
Like some initial trees in XTAG \cite{tech-rept01}, such trees can have
a {\em pair} of anchors.
Since there are different ways in which discourse units can be
parallel, we assume a different initial tree for {\em contrast}
(``on the one hand''... ``on the other (hand)''...),
{\em disjunction} (``either''... ``or''...),
{\em addition} (``not only''... ``but also''...),
and {\em concession} (``admittedly''... ``but''...).

Finally, there are initial trees for structural connectives between
adjacent sentences or clauses that convey a particular relation
between the connected units. One clear example is ``so'', conveying
{\em result}. Its initial tree is shown in
Figure~\ref{relandor:fig}.
\begin{figure*}
\centerline{\epsfig{file=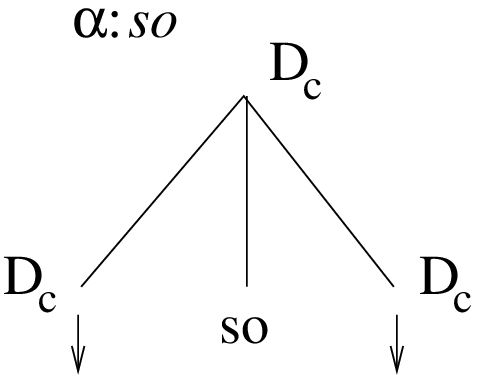, scale=0.60}} 
\caption{Initial tree for coordinate conjunction ``so''.}
\label{relandor:fig}
\end{figure*}
We will have a better sense of what other connectives
to treat as structural as a result of annotation efforts of the sort
described in \cite{creswell02}.\footnote{For example, one might also
  have initial trees for {\em marked} uses of ``and'' and
``or'', that have a specific meaning beyond simple conjunction or
disjunction as in
\eenumsentence{\label{marked:ex}
\item Throw another spit ball {\em and} you'll regret it.
\item Eat your spinach {\em or} you won't get dessert.}
These differ from the more frequent, simple coordinate uses of ``and''
and ``or'' in that the second conjunct in these marked cases bears a
discourse relation to the first conjunct ({\em result} in
both (\ref{marked:ex}a) and (\ref{marked:ex}b)). With simple
coordinate uses of ``and'' and ``or'', all conjuncts (disjuncts) bear
the same relation to the same immediately left-adjacent discourse unit.
For example, in (\ref{trustand:ex}), each
conjunct is a separate {\em explanation} for not trusting John, while in
(\ref{scope:ex}), each disjunct conveys an alternative result of John's
good fortune.
\enumsentence{You shouldn't trust John. He never returns what he
borrows, {\em and} he bad-mouths his associates behind their
backs.\label{trustand:ex}}  
\enumsentence{John just won the lottery. So he will quit his job, {\em or} he
will at least stop working overtime.\label{scope:ex}}
For simple coordinate uses of ``and'' and ``or'', we have auxiliary
trees (Section \ref{aux-tree:sec}).}

\subsubsection{Auxiliary Trees}
\label{aux-tree:sec}

DLTAG uses auxiliary trees in two ways: (a) for discourse
units that continue a description in some way; and (b) for discourse
adverbials. Again we describe each in turn. 

\begin{figure*}
\centerline{\epsfig{file=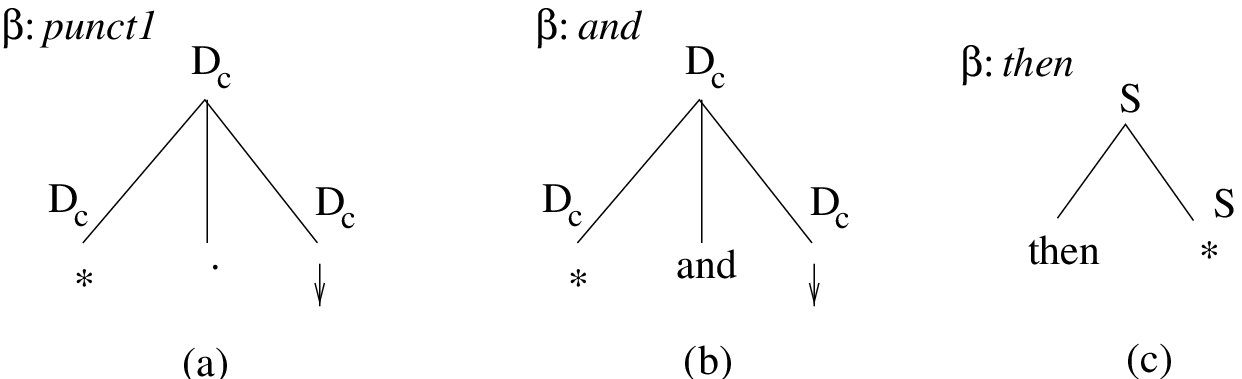,scale=.75}}
\caption{Auxiliary trees for basic elaboration. These particular trees
are anchored by (a) the punctuation mark ``.'' and (b) ``and''. The
symbol $*$ indicates the {\em foot node} of the auxiliary tree, which
has the same label as its root. (c) Auxiliary tree for the discourse
adverbial ``then''.}
\label{modtree:fig}
\end{figure*}

First, auxiliary trees anchored by punctuation (e.g. {\em period},
{\em comma}, {\em semi-colon}, etc.) (Figure~\ref{modtree:fig}a)
or by {\em simple coordination}  (Figure~\ref{modtree:fig}b) are used
to provide further description of a situation or of one or more
entities (objects, events, situations, states, etc.) within the
situation\footnote{The latter use of an auxiliary tree is related to
{\em dominant topic chaining} in \cite{scha88} and {\em entity chains}
in \cite{knott:be}.} The additional information is conveyed
by the discourse clause that fills its substitution site. Such
auxiliary trees are used in the derivation of simple discourses such
as: 
\enumsentence{
a. John went to the zoo.\\
b. He took his cell phone with him. \label{zoo1}
}

\begin{figure*}
\centerline{\epsfig{file=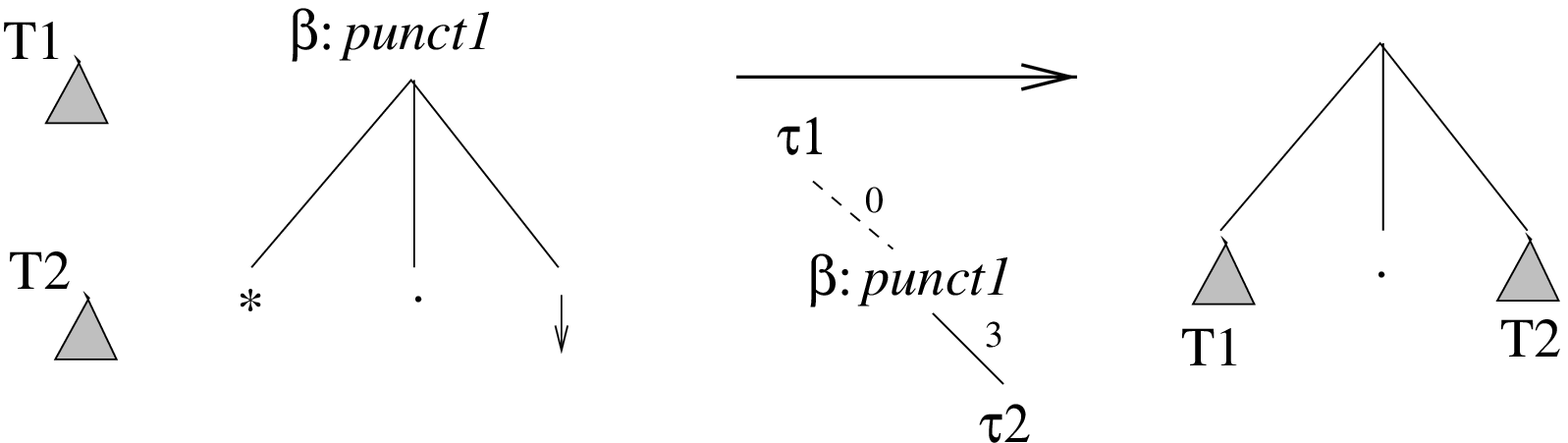,scale=.6}}
\caption{TAG derivation of Example~\ref{zoo1}}
\label{zoo1:fig}
\end{figure*}
Figure~\ref{zoo1:fig} shows the DLTAG derivation of
Example~\ref{zoo1}, starting from LTAG derivations of the individual
sentences.\footnote{We comment on left-to-right incremental
  construction of DLTAG structures in parallel with sentence-level
  LTAG structures at the end of Section~\ref{tagex:sec}.}
To the left of the arrow ($\rightarrow$) are the elementary trees to
be combined: T1 stands for the LTAG tree for clause \ref{zoo1}a, T2
for clause \ref{zoo1}b, and $\beta$:\emph{punct1}, for the auxiliary tree
assocated with the full stop after (\ref{zoo1}a).
In the derivation, the foot node of $\beta$:\emph{punct1} is adjoined
to the root of T1 and its substitution site filled by T2, resulting in
the tree to the right of $\rightarrow$. (A standard way of 
indicating TAG {\em derivations} is shown under $\rightarrow$, where dashed
lines indicate adjunction, and solid lines, substitution. Each
line is labelled with the address of the argument at which the
operation occurs. $\tau 1$ is the derivation tree for T1, and  $\tau
2$, the derivation tree for T2.)

The other auxiliary trees used in the lexicalised discourse grammar
are those for discourse adverbials, which are simply
auxiliary trees in a sentence-level LTAG \cite{tech-rept01}, but with
an interpretation that projects up to the discourse level. An
example is shown in Figure~\ref{modtree:fig}c. Adjoining such an
adverbial to a clausal/sentential structure contributes to how
information conveyed by that structure relates to the previous
discourse. 

There is some lexical ambiguity in this grammar, but no more than
serious consideration of adverbials and conjunctions demands.
First, as already noted, discourse adverbials have other uses that
may not be anaphoric (\ref{non-ac:ex}a-b) and may not be clausal
(\ref{non-ac:ex}a-c):
\enumsentence{a. John ate an apple {\em instead} of a pear.\\
b. {\em In contrast} with Sue, Fred was tired.\\ 
c. Mary was {\em otherwise} occupied.\label{non-ac:ex}}
Secondly, many of the adverbials found in second position in
parallel constructions (e.g., ``on the other hand'', ``at the same
time'', ``nevertheless'') can also serve as simple adverbial discourse
connectives on their own. In the first case, they will be one of the
two anchors of an initial tree (Figure~\ref{par:fig}), while in the
second, they will anchor a simple auxiliary tree
(Figure~\ref{modtree:fig}c). These lexical ambiguities correlate with
structural ambiguity. 

\subsection{Example Derivations}
\label{tagex:sec}

It should be clear by now that our approach aims to explain
discourse semantics in terms of a product of the same three
interpretive mechanisms that operate within clause-level semantics:
\begin{itemize}
\addtolength{\itemsep}{-4pt}
\item compositional rules on syntactic structure (here, discourse structure)
\item anaphor resolution
\item inference triggered by adjacency and structural connection.
\end{itemize}
For the compositional part of semantics in DLTAG (in particular,
computing interpretations on derivation trees), we follow
\namecite{jv99}. Roughly, they compute interpretations on the derivation
tree by a bottom-up procedure. At each level, function-application is
used to assemble the interpretation of the tree from the
interpretation of its root node and its subtrees.  Where multiple 
subtrees have function types, the interpretation procedure is
potentially nondeterministic: The resulting ambiguities in
interpretation may be admitted as genuine, or they may be eliminated
by a lexical specification. Multi-component TAG tree-sets are used to
provide an appropriate compositional treatment for quantifiers, which
we borrow for interpretating ``for example'' (Examples~\ref{trust:ex}c-d).

We show here rather informally how DLTAG and an interpretative
process on its derivations operate. We start with previous examples
(\ref{trustbecause:ex}) (here \ref{trust:ex}c) and (\ref{trustrep:ex})
(here, \ref{trust:ex}d) and two somewhat 
simpler variants (\ref{trust:ex}a-b): 
\eenumsentence{\label{trust:ex}
\item You shouldn't trust John because he never returns what he
borrows.
\item You shouldn't trust John. He never returns what he borrows.
\item You shouldn't trust John because, for example, he never
returns what he borrows.
\item You shouldn't trust John. For example, he never returns what he
borrows.}
This will allow us to show how (\ref{trust:ex}a-b) and
(\ref{trust:ex}c-d) receive similar interpretations, despite having
somewhat different derivations, and how the discourse adverbial ``for
example'' contributes both syntactically and semantically to
those interpretations.

We let  $T1$ stand for the LTAG parse tree for ``you shouldn't trust
John'', $\tau 1$, its derivation tree, and interp($T1$), the
eventuality associated with its interpretation. Similarly, we let $T2$
stand for the LTAG parse tree for ``he never returns what he borrows'',
$\tau 2$, its derivation tree, and interp($T2$), the eventuality
associated with its interpretation.

\begin{figure*}
\centerline{\epsfig{file=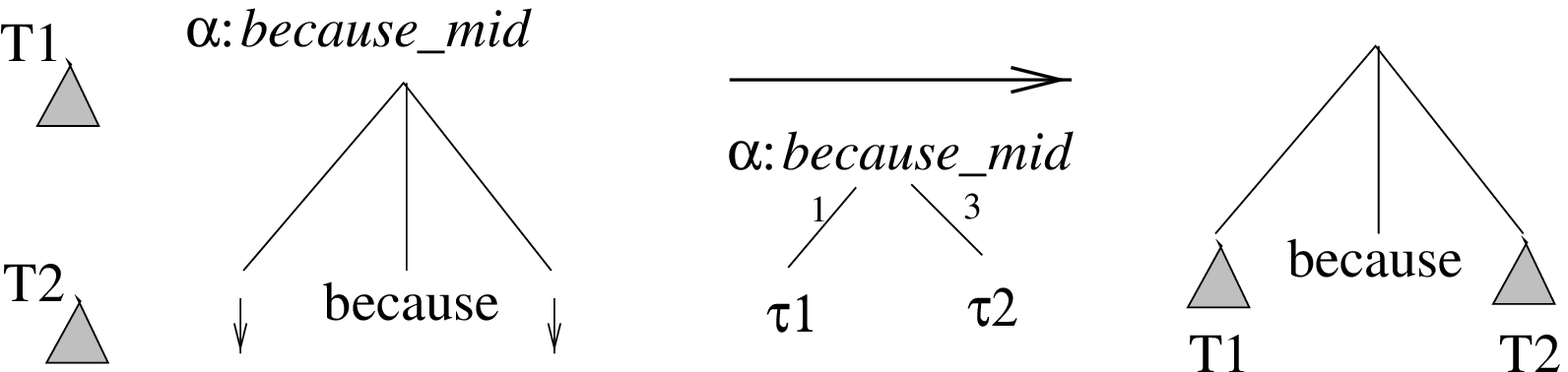,scale=0.6}}
\caption{Derivation of Example~\ref{trust:ex}a. The derivation tree is
shown below the arrow, and the derived tree, to its right. (Node
labels D$_c$ have been omitted for simplicity.)}
\label{minPa:fig}
\end{figure*}

Example~\ref{trust:ex}a involves an initial tree 
($\alpha$:{\em because-mid}) anchored by ``because''
(Figure~\ref{minPa:fig}). Its derived tree comes from $T1$
substituting at the left-hand substitution site of 
$\alpha$:{\em because-mid} (index 1) and $T2$ at its right-hand
substitution site (index 3). Compositional interpretation of
the resulting derivation tree yields
{\em explanation}(interp($T2$),interp($T1$)). (A more  
precise interpretation would distinguish between the direct and
epistemic causality senses of ``because'', but the derivation would
proceed in the same way.) 

\begin{figure*}
\centerline{\epsfig{file=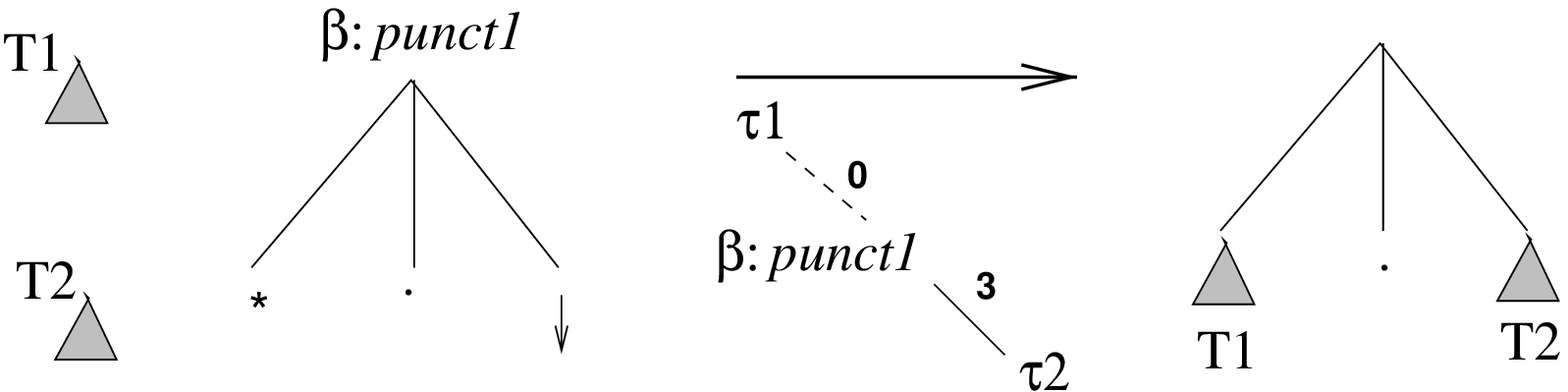,scale=0.6}}
\caption{Derivation of Example~\ref{trust:ex}b}
\label{minPb:fig}
\end{figure*}

In contrast with (\ref{trust:ex}a), Example~\ref{trust:ex}b employs an
auxiliary tree ($\beta$:{\em punct1}) anchored by full-stop ``.''
(Figure~\ref{minPb:fig}). Its derived tree comes from $T2$
substituting at the right-hand substitution site (index 3) of
$\beta$:{\em punct1}, and $\beta$:{\em punct1} adjoining at the root
of $T1$ (index 0). Compositional interpretation of the
derivation tree yields merely that $T2$ continues the description of
the situation associated with $T1$ -- i.e., {\em
elaboration}(interp($T2$),interp($T1$)). Further inference 
triggered by adjacency and structural connection leads to a conclusion
of causality between them -- i.e.,
{\em explanation}(interp($T2$),interp($T1$)), but this conclusion is
defeasible because it can be denied without a contradiction -- e.g. 
\enumsentence{You shouldn't trust John. He never returns what he borrows.
But that's not why you shouldn't trust him.}

\begin{figure*}
\centerline{\epsfig{file=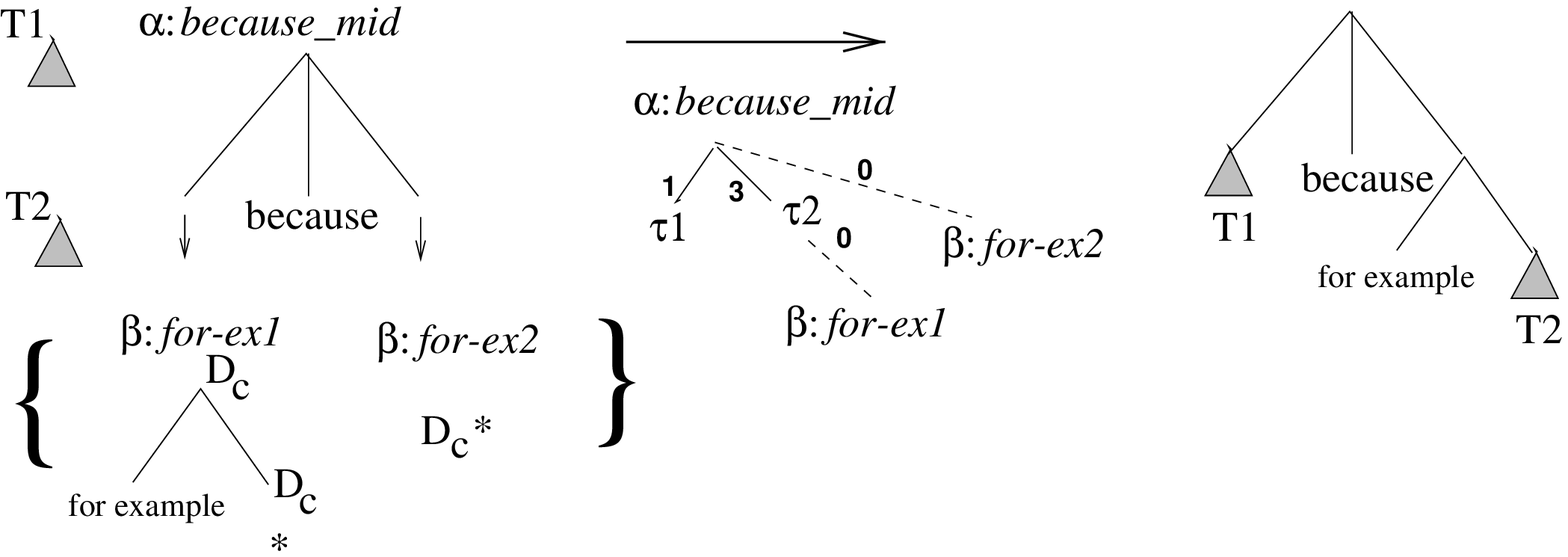,scale=0.6}}
\caption{Derivation of Example~\ref{trust:ex}c}
\label{minPc:fig}
\end{figure*}

Example~\ref{trust:ex}c differs from (\ref{trust:ex}a) in containing
``for example'' in its second clause. As noted earlier, ``for
example'' resembles a quantifier with respect to its semantics, as its
interpretation takes wider scope than would be explained by its
syntactic position. We handle this in the same way that quantifiers are
handled in \cite{jv99}, by associating with ``for example'' a
two-element TAG tree-set  (Figure~\ref{minPc:fig}). Both trees in the
tree-set participate in the derivation: the auxiliary tree
$\beta$:{\em for\_ex1} adjoins at the root of $T2$, while the
auxiliary tree  $\beta$:{\em for\_ex2} adjoins at the root of the
higher discourse unit. Since we saw from Example~\ref{trust:ex}a that
the interpretation of this higher discourse unit is
{\em explanation}(interp($T2$),interp($T1$)), the interpretation
associated with the adjoined $\beta$:{\em for\_ex2} node both embeds
and abstracts this interpretation, yielding
\begin{quote}
{\em exemplification}(interp($T2$), $\lambda$X . {\em
   explanation}(X,interp($T1$))
\end{quote}
That is, John's never returning what he borrows is one instance of a set
of explanations.

\begin{figure*}
\centerline{\epsfig{file=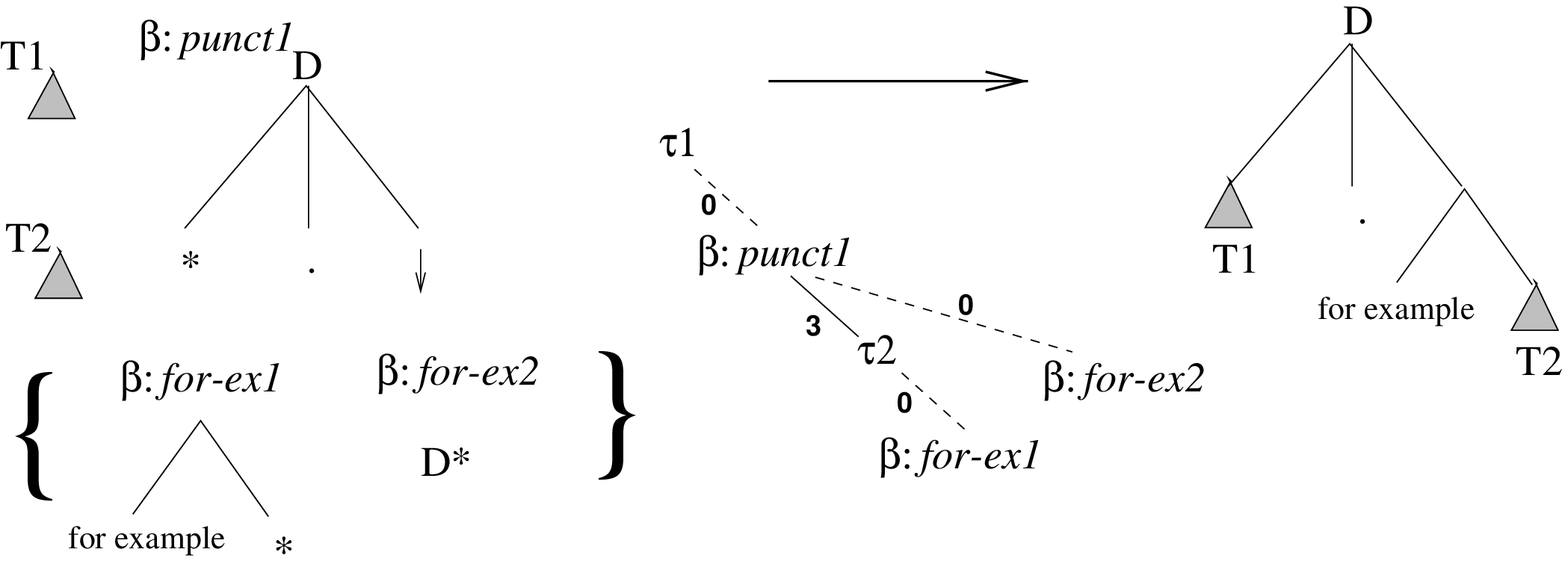,scale=0.6}}
\caption{Derivation of Example~\ref{trust:ex}d}
\label{minPd:fig}
\end{figure*}

Similarly, Example~\ref{trust:ex}d differ from (\ref{trust:ex}b) in
containing ``for example'' in its second sentence.
As in Example~\ref{trust:ex}b, an inferred relation is triggered 
between the interpretations of $T2$ and $T1$, namely
{\em explanation}(interp($T2$),interp($T1$)). Then, as a result of
$\beta$:{\em for\_ex1} adjoining at $T2$ and $\beta$:{\em for\_ex2}
adjoining at the root of the higher discourse unit,
``for example'' again contributes the interpretation
\begin{quote}
{\em exemplification}(interp($T2$), $\lambda$X . {\em
   explanation}(X,interp($T1$))  
\end{quote}
Thus (\ref{trust:ex}c) and  (\ref{trust:ex}d) only differ in the
derivation of the interpretation that ``for example'' then abstracts
over.

The next example we will walk through is Example~\ref{barolo:ex}
(given here as Example~\ref{barolo:rep}).
\enumsentence{
John loves Barolo. So he ordered three cases of the '97.
But he had to cancel the order because {\em then} he discovered he was
broke.\label{barolo:rep}} 
As shown in Figure~\ref{baroldev:fig}, this example involves two
initial trees ($\alpha$:{\em so}, $\alpha$:{\em because\_mid}) for the
structural connectives ``so'' and ``because''; an auxiliary tree for
the structural connective ``but'' ($\beta$:{\em but}), since ``but''
functions as a simple conjunction to continue the description of the
situation under discussion; an auxiliary tree ($\beta$:{\em then}) for
the discourse adverbial ``then''; and initial trees for the
four individual clauses $T1$-$T4$. As can be seen from the derivation
tree, $T1$ and $T2$ substitute into
$\alpha$:{\em so} as its first and third arguments, and
$\beta$:{\em but} root-adjoins to the result. The substitution
argument of $\beta$:{\em but} is filled by
$\alpha$:{\em because\_mid}, with $T3$ and $T4$ substituted in as its
first and third arguments, and $\beta$:{\em then} is root-adjoined to $T4$.
The interpretation contributed by ``then'', after its
anaphoric argument is resolved to interp($T2$), is 
\begin{quote}
$\iota$4: {\em after}(interp($T4$), interp($T2$)).
\end{quote}
The interpretations derived compositionally from the structural
connectives ``so'', ``because'' and ``but'' are:
\begin{quote}
$\iota$1: {\em result}(interp($T2$), interp($T1$))\\
$\iota$2: {\em explanation}(interp($T4$), interp($S3$))\\
$\iota$3: {\em elaboration}($\iota$2,$\iota$1)
\end{quote}
Further inference may then refine {\em elaboration} to {\em contrast},
based on how but is being used.

\begin{figure*}
\centerline{\epsfig{file=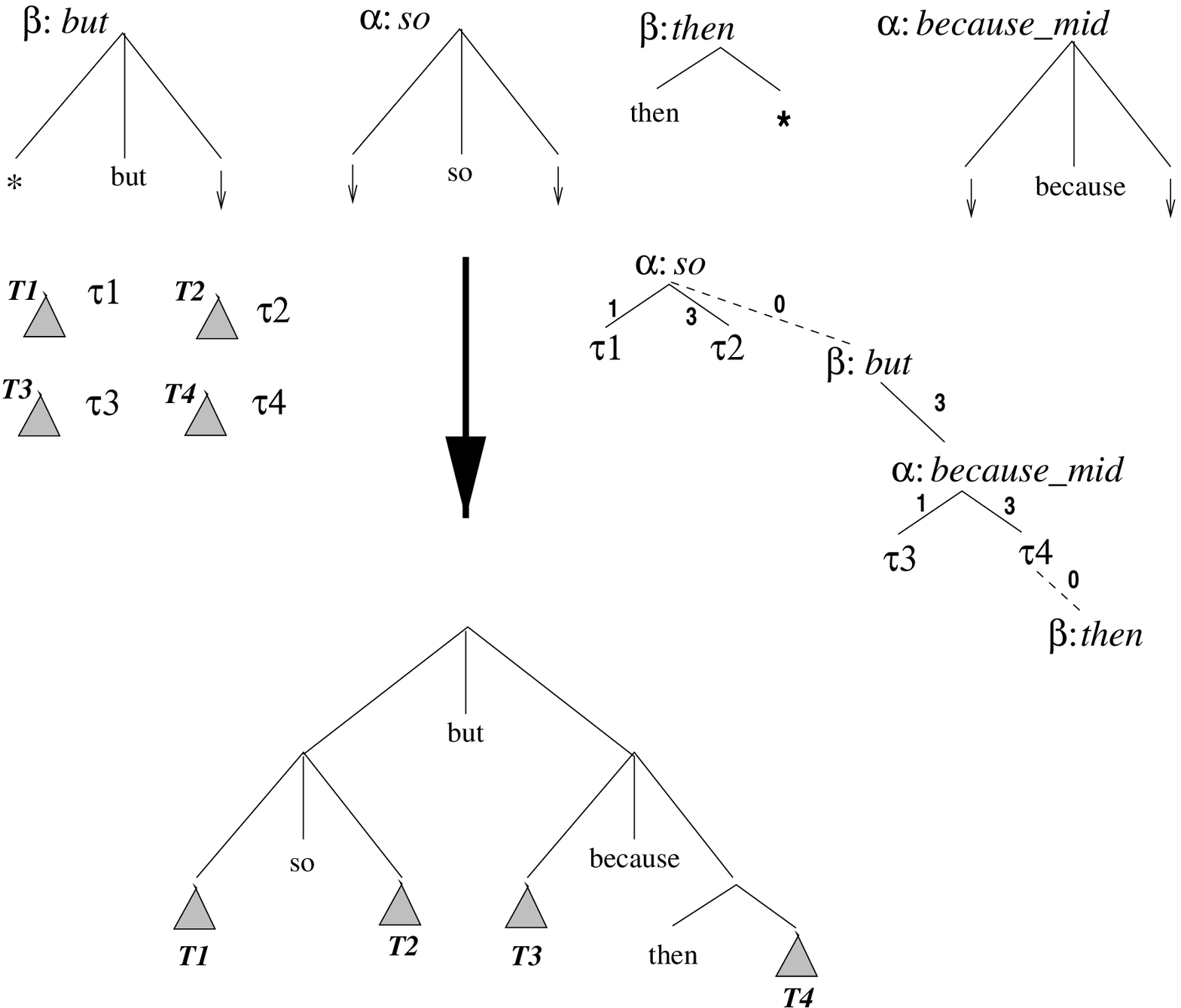,scale=0.5}}
\caption{Derivation of Example~\ref{barolo:rep}}
\label{baroldev:fig}
\end{figure*}

Finally, we want to point out one more way in which texts that seem
to be close paraphrases get their interpretations in different
ways. Consider the two texts in Example~\ref{this-because:ex}:
\eenumsentence{\label{this-because:ex}
\item You should eliminate part2 before part3
because part2 is more susceptible to damage.
\item  You should eliminate part2 before part3. This is
because part2 is more susceptible to damage.}
Example~\ref{this-because:ex}b is a simpler version of an example in
\cite{mm95}, where ``This is because'' is treated as an unanalyzed cue
phrase, no different from ``because'' in (\ref{this-because:ex}a). We
show here that this isn't necessary: One can analyze
(\ref{this-because:ex}b) using compositional semantics and anaphor
resolution, and achieve the same results. 

\begin{figure*}
\centerline{\epsfig{file=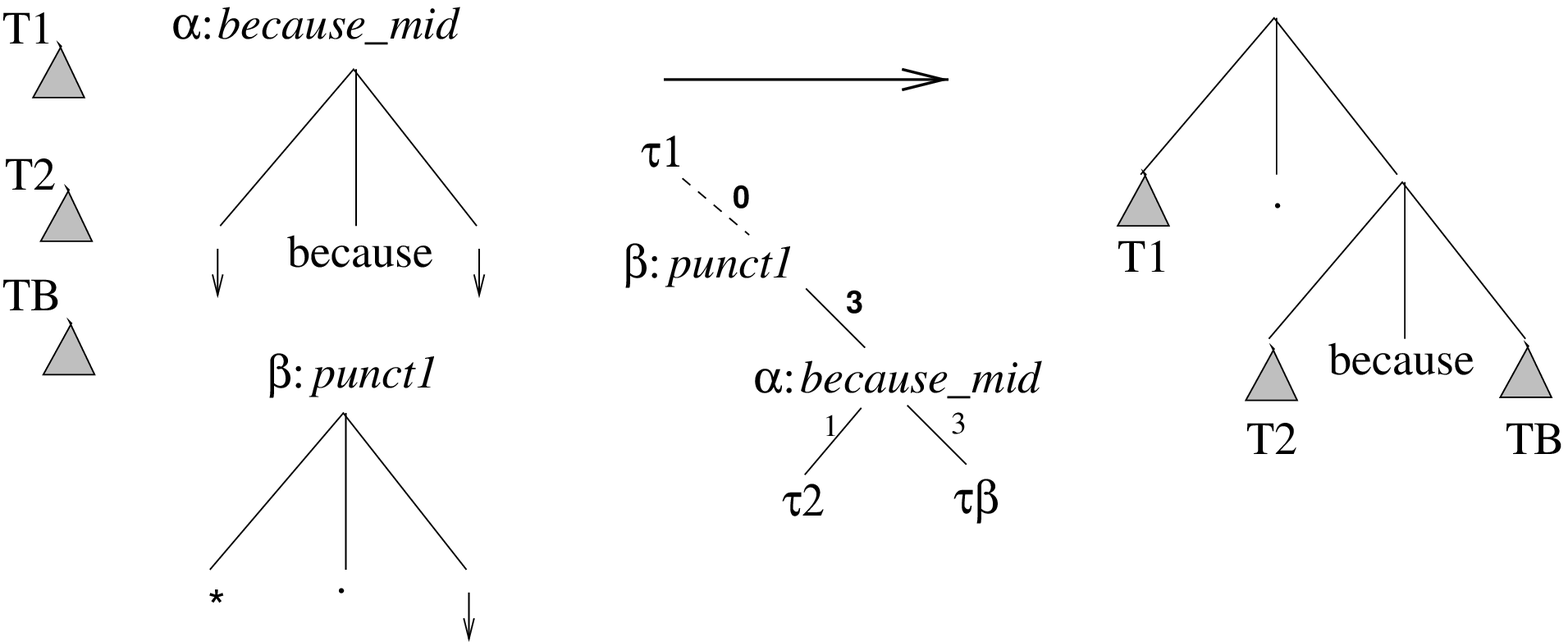,scale=0.6}}
\caption{Derivation of Example~\ref{this-because:ex}b}
\label{this-because:fig}
\end{figure*}

First consider (\ref{this-because:ex}a). Given the interpretations of
its two component clauses, its overall interpretation follows in the same
way as (\ref{trust:ex}a), shown in
Figure~\ref{minPa:fig}. Now consider (\ref{this-because:ex}b) and the
derivation shown in Figure~\ref{this-because:fig}.
Here the initial tree $\alpha$:{\em because-mid} has its two arguments
filled by $T2$, the TAG analysis of ``this is'' and $TB$, the TAG analysis
of ``part 2 is more susceptible to damage''. The overall derived tree
for (\ref{this-because:ex}b) comes from $\beta$:{\em punct1}
root-adjoining to $T1$ (the TAG analysis of ``You should
eliminate part2 before part3''), with the subsitution site of  
$\beta$:{\em punct1} filled by the $\alpha$:{\em because-mid}
derivation. The compositional interpretation of the derivation tree
yields the interpretation of the $\alpha$:{\em because-mid} tree
($i1$) as an {\em elaboration} of the interpretation of $T1$:
\begin{quote}
$i1$: {\em explanation}(interp($TB$),interp($T2$))\\
$i2$: {\em elaboration}($i1$,interp($T1$))
\end{quote}
But this is not all. The pronoun ``this'' in $T2$ is resolved
anaphorically to the nearest consistent eventuality
\cite{eckert01,byron02}, which in this case is interp($T1$). Taking
this as the interpretation of $T2$ and substituting, we get
\begin{quote}
$i1$: {\em explanation}(interp($TB$),interp($T1$))\\
$i2$: {\em elaboration}($i1$,interp($T1$))
\end{quote}
Notice that $i1$ is also the interpretation of
(\ref{this-because:ex}a). To this, $i2$ adds the somewhat redundant
information that $i1$ serves to {\em elaborate} the advice in
$T1$. Thus (\ref{this-because:ex}a) and (\ref{this-because:ex}b)
receive similar interpretations but by different means. This
treatment has the added advantage that one does not have to treat
``This is not because'' as a separate cue phrase. Rather, negation
simply produces
\begin{quote}
$i1$: $\neg${\em explanation}(interp($TB$),interp($T1$))\\
$i2$: {\em elaboration}($i1$,interp($T1$))
\end{quote}
That is, $T1$ is elaborated by a denial of a (possible)
explanation. Presumably, the text would go on to provide the actual
explanation.

Finally, we want to comment on the holy grail of discourse parsing:
running it in parallel with incremental sentence-level
parsing. Neither the analyses given in this section, nor the discourse
parser described in \cite{dltag01} run in parallel with incremental
sentence-level parsing. But we believe that an approach grounded in a
lexicalized grammar holds more promise for parallel, incremental
sentence-discourse processing than either an approach that uses
distinct mechanisms for the two, or an approach that uses
phrase-structure rules for both.

An approach to sentence-discourse processing that was both incremental
and parallel would minimally require the following:
\begin{itemize}
\item A left-to-right parser for the lexicalized grammar that would
  simultaneously compute increments to both sentence-level syntactic
  structure, sentence-level semantics, discourse-level syntactic
  structure and discourse-level semantics. Increments to the latter
  two would only occur at clause boundaries and with discourse
  adverbials and structural connectives.

\item An incremental anaphor resolution mechanism, similar to that in
  \cite{strube98}, but extended both to deictic pronouns, as in
  \cite{eckert01,byron02}, and to the anaphoric argument of discourse
  adverbials. 

\item Incremental computation of discourse structure in terms of {\em
    elaboration} relations and further non-defeasible reasoning to
  more specific relations, where possible.
\end{itemize}

An left-to-right parser that simultaneously produces sentence-level
syntactic and 
semantic analyses already exists for combinatory categorial grammar
\cite{steed96,steed00,hbb02}, and it would seem straight-forward to extend
such a parser to computing discourse-level syntax and
semantics as well. Similarly, it seems straight-forward to produce an
incremental version of any of the current generation of anaphor
resolution mechanisms, extended to deictic pronouns, although current
approaches only attempt to resolve ``this'' and ``that'' with
the interpretation of a single clause -- not with that of any larger
discourse unit. As these approaches are also not very accurate as yet,
incremental anaphor resolution awaits improvements to anaphor
resolution in general. Moreover, as we better understand the specific
anaphoric properties of discourse adverbials through empirical
analysis such as \cite{creswell02}, such anaphor 
resolution mechanisms can be extended to include them as well.

As for building discourse structure incrementally in parallel with
syntactic structure, there is no working prototype yet that will do
what is needed, but we have no doubt that better understanding of
semantics and researchers' reliable ingenuity will eventually succeed
here as well.

\section{Conclusion}

In this paper, we have argued that discourse adverbials make an
anaphoric, rather than a structural, connection with the previous
discourse (Section~\ref{anaph:sec}), and we have provided a general
view of anaphora in which it makes sense to talk of discourse
adverbials as being anaphoric (Section~\ref{frame:sec}). We have then
shown that this view of discourse adverbials allows us to characterize
a range of ways in which the relation contributed by a discourse
adverbial can interact with the relation conveyed by a structural
connective or inferred through adjacency (Section~\ref{inf:sec}),
and then shown how discourse syntax and semantics can be treated as an
extension of sentence-level syntax and semantics, using a lexicalised
discourse grammar (Section~\ref{tag:sec}).

We are clearly not the first to have proposed a
grammatical treatment of low-level aspects of discourse semantics
\cite{al99,gardent97,pol-van96,scha88,schilder97b,schilder97a,vandenberg96}.
But we are the first to have recognised that a key to avoiding
problems of maintaining a compositional
semantics for discourse lies in recognizing discourse adverbials 
as anaphors and not trying to shoe-horn everything into a single class
of discourse connectives. While we are not yet able to propose a
solution to the problem of correctly resolving discourse adverbials or
a way of achieving the holy grail of computing discourse syntax and
semantics in parallel with incremental sentence processing,
the proposed approach does simplify issues of discourse structure and
discourse semantics in ways that have not before been possible.

\starttwocolumn
\begin{acknowledgments}
The authors would like to thank Kate Forbes, Katja Markert, Natalia
Modjeska, Rashmi Prasad, Eleni Miltsakaki, Cassandra Creswell, Mark
Steedman, members of the University of Edinburgh Dialogue Systems
Group, and participants at ESSLLI'01, for 
helpful criticism as the ideas in the paper were being developed.
We would also like to thank our three anonymous reviewers. We believe
that in addressing their criticisms and suggestions, both the
paper's arguments and its presentation have become clearer.
\end{acknowledgments} 

\nocite{*}

\bibliographystyle{fullname}

\bibliography{webber-01}
\end{document}